\documentclass[11pt]{article}

\usepackage[preprint]{acl}

\usepackage{times}
\usepackage{latexsym}

\usepackage[T1]{fontenc}

\usepackage[utf8]{inputenc}

\usepackage{microtype}

\usepackage{inconsolata}

\usepackage{graphicx}
\usepackage{booktabs}
\usepackage{array}
\usepackage{multirow}
\usepackage{amsmath}
\usepackage{amssymb}
\usepackage{fvextra}
\usepackage[most]{tcolorbox}
\usepackage{xcolor}
\usepackage{listings}
\usepackage{tabularx}
\usepackage{hyperref}
\usepackage{url}
\usepackage{float}
\usepackage{afterpage}
\newcolumntype{Y}{>{\raggedright\arraybackslash}X}

\usepackage{subcaption}
\usepackage{tablefootnote}
\usepackage{enumitem}

\lstdefinestyle{promptstyle}{
  basicstyle=\ttfamily\scriptsize,
  breaklines=true,
  breakatwhitespace=false,
  columns=fullflexible,
  keepspaces=true,
  showstringspaces=false,
  upquote=true,
  tabsize=2
}

\newtcblisting{promptbox}[1]{
  enhanced,
  listing only,
  colback=gray!6,
  colframe=gray!45,
  title={#1},
  fonttitle=\bfseries\small,
  coltitle=black,
  boxrule=0.4pt,
  arc=1.5mm,
  left=1mm,
  right=1mm,
  top=1mm,
  bottom=1mm,
  listing options={style=promptstyle},
  breakable
}

\title{LitSeg: Narrative-Aware Document Segmentation for Literary RAG}

\author{Ruikang Zhang \quad Zhanni Chen\thanks{\quad Equal contribution.} \quad Yiqiao Cai\footnotemark[1] \quad Qi Su\thanks{\quad Corresponding author.} \\
        Peking University, Beijing, China \\
        \texttt{\{2300018416, 2300018109, 2300018316\}@stu.pku.edu.cn}, \texttt{sukia@pku.edu.cn}}

\begin{document}
\maketitle
\begin{abstract}
Retrieval-Augmented Generation (RAG) enhances Large Language Models (LLMs) by incorporating external knowledge, particularly for long-tail domains such as literary works. However, the critical step of document segmentation in RAG remains largely underexplored. Existing strategies typically either ignore semantics or overlook the complicated narrative structures of literary works, often resulting in chunks with fragmented plots and unclear references that hinder retrieval and generation performance.
To address this, we propose \textbf{LitSeg}, a novel narrative-theory-guided segmentation framework. By employing multi-stage prompting, LitSeg explicitly extracts valid events, clarifies narrative structures, and locates turning points to inform segmentation.
To alleviate the computational overhead of multi-stage inference with large-scale models, we further introduce \textbf{LitSeg-Lite}, a lightweight single-pass chunker fine-tuned on LitSeg-generated data via a two-stage training strategy, distilling the complex process into a single inference pass.
Extensive experiments demonstrate that compared to baselines, our methods yield high-quality text chunks that are narratologically coherent and self-contained. This improved segmentation quality enhances retrieval accuracy and context relevance, and boosts downstream QA performance. Ablation studies validate the efficacy of narratological guidance and data distillation, and efficiency analysis shows that LitSeg-Lite matches the teacher at a substantially lower inference cost.
\end{abstract}

\section{Introduction}

\begin{figure*}
    \centering
    \includegraphics[width=1\linewidth]{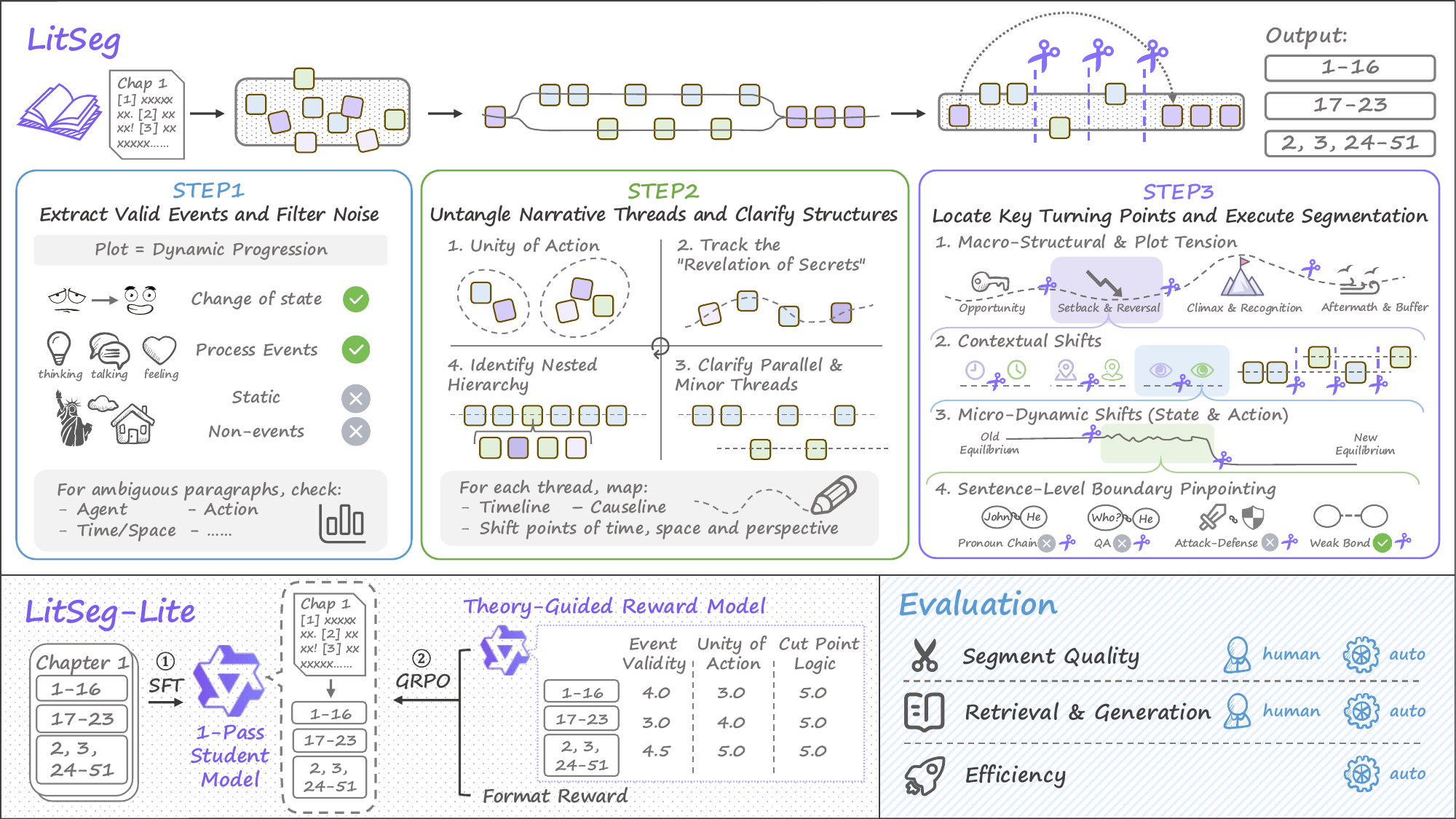}
    \caption{Overview of our proposed framework. \textbf{LitSeg} leverages narratological theories to guide a high-capacity teacher model through a multi-stage segmentation pipeline, producing narrative-aware chunking annotations. \textbf{LitSeg-Lite} then distills this knowledge into a lightweight student model via SFT and GRPO with theory-guided rewards. LitSeg and LitSeg-Lite produce high-quality chunks that are narratologically cohesive and self-sufficient, facilitating effective RAG retrieval and generation.}
    \label{fig:overview}
\end{figure*}

Retrieval-Augmented Generation (RAG) enhances Large Language Models (LLMs) by retrieving relevant chunks from external corpora, thereby improving their access to long-tail knowledge such as literary works~\cite{Zhao2026RAG}.
An RAG system typically consists of three parts: 1) indexing, in which raw documents are segmented into chunks, and then represented and stored in vector databases; 2) retrieval, in which the system retrieves relevant chunks given a user query; 3) generation, in which an LLM formulates a response given the query and retrieved chunks~\cite{11300411}.
In such systems, the semantic integrity and coherence of document chunks directly dictate the quality of vector representation and available context, thereby influencing retrieval accuracy and the performance of RAG systems on downstream tasks such as question answering (QA)~\cite{10.1007/978-3-032-15404-0_30} and agentic tasks~\cite{singh2025agenticretrievalaugmentedgenerationsurvey}.
However, existing work has primarily focused on index structures, document retrieval, and response generation~\citep[e.g.,][]{huang-etal-2025-retrieval,gao-etal-2023-precise,10.1145/3726302.3730078}, leaving document segmentation underexplored~\cite{wang-etal-2025-document}.
Given the complicated narrative structure of literary works~\cite{ShenWang2010}, inappropriate text segmentation can distort the meaning of individual chunks and lead to fragmented plots and unclear references that degrade downstream retrieval and generation performance.
Nevertheless, existing segmentation strategies either ignore semantics or overlook such high-order structures (see Section~\ref{subsec:seg-in-rag}), leaving this critical issue unresolved.

In light of this, we present \textbf{LitSeg} (Section~\ref{subsec:litseg}), a narrative-theory-guided segmentation framework.
Drawing on established theories, we leverage explicit, multi-stage prompting to task an LLM with extracting valid events, clarifying narrative structures, locating key turning points, and finally executing segmentation.
We also provide the model with long context and design an input-output format that numbers the input sentences and represents each chunk as a main sentence range with optional context indices, thereby supporting flexible incorporation of non-adjacent sentences to preserve broader narrative context while representing chunks with low token overhead. %
To further reduce the overhead of multi-stage inference and large-scale model usage, we also present \textbf{LitSeg-Lite} (Section~\ref{subsec:litseg-lite}), a lightweight chunker that performs segmentation in a single inference pass.
Fine-tuned on LitSeg-generated data via a two-stage training strategy, it achieves performance comparable to LitSeg.
Experiments (Section~\ref{sec:exp}) show that compared to baselines, LitSeg and LitSeg-Lite produce high-quality, narratively cohesive, and self-contained text chunks, improving retrieval and generation performance.
Ablation studies further show that both narrative theories and distillation of LitSeg data contribute to the strong performance of LitSeg-Lite.
Efficiency analysis demonstrates that LitSeg-Lite matches LitSeg and outperforms strong LLM-based baselines at substantially lower inference cost.
Figure~\ref{fig:overview} provides an overview of our proposed framework. 

Our main contributions are threefold:

\begin{enumerate}[wide, nosep, itemsep=0pt, topsep=0pt, partopsep=0pt, labelindent=0pt, leftmargin=*]
    \item We propose \textbf{LitSeg}, a novel narrative-theory-guided segmentation framework for literary works. By employing multi-stage prompting, LitSeg explicitly extracts narrative structure to aid segmentation, effectively addressing the lack of semantic awareness of existing segmentation methods and preserving high-order narrative structures, while representing chunks via sentence indices allows incorporating non-adjacent sentences, preserving long-range semantics with reduced token usage.
    \item We introduce \textbf{LitSeg-Lite}, a lightweight, single-pass chunker fine-tuned on LitSeg-generated data via a two-stage training strategy. It distills the multi-step inference process into a single inference pass, achieving performance comparable to LitSeg with substantially lower inference overhead.
    \item Comprehensive experiments demonstrate that our methods produce high-quality text chunks that are narratologically coherent and self-contained, improving retrieval and downstream QA performance. Ablation studies validate the efficacy of narratological guidance and data distillation. Efficiency analysis confirms the practical advantages of LitSeg-Lite over existing LLM-based approaches.
\end{enumerate}

\section{Related Work}
\label{sec:rw}

\subsection{Document Segmentation in RAG}
\label{subsec:seg-in-rag}

In RAG systems, the most prevalent segmentation methods typically split documents by fixed token~\cite{11300411} or character counts.
Recursive character splitters~\cite{Chase_LangChain_2022} refine this strategy by recursively applying common delimiters (e.g., newlines) until size constraints are met.
However, these heuristic approaches are semantically agnostic, often disrupting structural integrity and semantic coherence.
To address this, semantics-aware methods have emerged, leveraging embedding models or LLMs to capture textual meaning.
Embedding-based approaches detect semantic shifts; for instance, SemanticChunker~\cite{Chase_LangChain_2022} inserts boundaries where the cosine distance between adjacent sentence-group embeddings exceeds a percentile threshold.
LLM-based methods further exploit model internals or reasoning capabilities: Perplexity Chunker~\cite{zhao2025metachunkinglearningtextsegmentation} identifies boundaries at local minima of the sentence-level loss curve; 
Margin-Sampling Chunker~\cite{zhao2025metachunkinglearningtextsegmentation} frames boundary detection as a binary decision based on the LLM's probability margin for keeping adjacent text; and LumberChunker~\cite{duarte-etal-2024-lumberchunker} uses a sliding window to prompt an LLM to directly pinpoint content shifts.
However, these methods still overlook the high-order narrative structures of literary works, where coherent chunks often depend on long-range causal relations, parallel narrative threads, and shifts in narrative level or perspective. This limitation restricts their effectiveness in literary RAG systems.

\subsection{Distillation for Efficient Task-Specific Models}
\label{subsec:distil}

Supervised Fine-Tuning (SFT) \cite{10.1145/3777411} remains the foundational step to adapt Large Language Models (LLMs) to expert demonstrations. While SFT effectively instills specific task knowledge and stylistic consistency, it is inherently constrained by the quality and diversity of the offline dataset \cite{chu2025sftmemorizesrlgeneralizes}. 
To address these limitations, Group Relative Policy Optimization (GRPO) \cite{shao2024deepseekmathpushinglimitsmathematical} has emerged as an efficient Reinforcement Learning (RL) paradigm.
GRPO generates a group of $G$ rollouts $\{y_1, y_2, \dots, y_G\}$ for each prompt $x$ and computes the relative advantage of each output by normalizing rewards within the group. This mechanism bypasses the optimization instabilities often associated with value function estimation and significantly reduces the memory overhead of traditional Proximal Policy Optimization (PPO) methods \cite{schulman2017proximalpolicyoptimizationalgorithms} that rely on a computationally expensive critic model to estimate value baselines.
Building upon the group-based efficiency of GRPO, Decoupled Clip and Dynamic Sampling Policy Optimization (DAPO) \cite{yu2025dapoopensourcellmreinforcement} further enhances training stability and output diversity, particularly in scenarios requiring long-sequence structural consistency.

\subsection{Narrative Theory and NLP}
\label{subsec:narrative-theory-and-nlp}

Narrative theory is the foundational framework for understanding how stories are constructed, structured, and communicated \cite{liveley2019narratology}, providing crucial theoretical underpinnings for various Natural Language Processing (NLP) tasks.

In the realm of narrativity analysis, narrative theory provides essential criteria for quantifying the degree of narrativity and distinguishing narrative discourse from non-narrative noise.
\citet{piper2021narrative} synthesizes classical and postclassical narrative theories, guiding computational narrative research by defining the core elements of narrativity. 
To operationalize event narrativity, \citet{gius2022towards} proposes an event-based plot model classifying textual components into distinct event types; \citet{piper2024using} further applies LLMs to annotate narrative discourse within the framework of \citet{genette1980narrative}'s narrative triangle concerning story, discourse, and narrating.

Beyond identifying narration, narrative theory also offers vital structural paradigms for organizing scattered events into coherent narrative threads, thereby facilitating storyline extraction. \citet{vossen2021narratology} introduces a narratology-based framework for news that maps the theoretical notions of \textit{fabula}, plot, and plot structure to explicit data structures: timelines, causelines, and storylines. To address the higher levels of non-linearity in fictional works compared to news, \citet{visser2025event} constructs a theoretical model for event annotation grounded in the theory of \textit{syuzhet} (the concrete presentation order of events) to achieve its computational detection.

On the other hand, narrative theory empowers text segmentation by guiding the identification of logical breakpoints across different levels. At a macro-structural level, \citet{papalampidi2019movie} models movie plots by identifying theory-grounded narrative turning points. At a micro-linguistic level, text boundaries are effectively pinpointed using mutual information theory: \citet{wagner2024automatic} detects zones of low Point-wise Mutual Information (PMI) between adjacent sentences to segment spoken testimonies, while \citet{wang2023m3seg} introduces M3Seg, a novel Maximum-Minimum Mutual information paradigm, to partition ASR transcripts.

Despite significant strides in quantifying narrative structures, these works are hindered by two primary limitations: theoretically, they often focus on single-faceted features, lacking a refined and synthesized comprehensive theoretical framework; practically, they predominantly target isolated computational tasks, thereby failing to induce and execute an end-to-end theoretical pipeline for the entire text segmentation process.

\section{Methodology}
\label{sec:methodology}

\subsection{LitSeg: Narrative-Theory-Guided Segmentation Framework}
\label{subsec:litseg}

\subsubsection{Theoretical Framework}

Our framework integrates classical and post-classical narratology with computational narratology, translating abstract literary concepts into operational metrics to address the limitations of existing semantic segmenters. Specifically, LitSeg explicitly decomposes the complex cognitive task of document chunking into a three-step narratological pipeline: Event Extraction, Thread Untangling, and Turning Points Pinpointing. See Appendix~\ref{app:prompt-litseg} for full prompts.

\paragraph{Extract Valid Events and Filter Noise}
Distinguishing valid narrative progression from background information is the foundational step of segmentation. 
Plot is defined by its dynamic, sequential nature; narrative elements only constitute a plot when they drive this dynamism \cite{scholes2006nature}. Thus, the event-based plot model \cite{gius2022towards} is adopted to explicitly task the LLM to filter text by retaining only changes of state and process events, while discarding static events and non-events (e.g., generic statements or counterfactual passages). 
For ambiguous paragraphs, the text's degree of narrativity is quantified by actively evaluating specific feature densities, such as the presence of an agent, sequential actions, spatial/temporal specifications, and rationale \cite{piper2021narrative}, to guarantee that only independent and valid events are extracted for downstream processing.

\paragraph{Untangle Narrative Threads and Clarify Structures}
Literary texts frequently feature intertwined storylines. To systematically disentangle these narratives, we adopt Aristotle's principle of unity of action \cite{butcher1902poetics,ShenWang2010}, enabling the model to precisely cluster events based on complete, core actions.
Since a well-arranged plot often builds the structural integrity of the whole story through the gradual unfolding of secrets \cite{forster1927aspects}, tracking this revelation process guides the model to connect scattered events into cohesive narrative chains. 
Furthermore, to handle complex discourse, the theory of simultaneity in narrative \cite{Margolin+2014+777+786} is incorporated to instruct the model to explicitly segment concurrent and parallel event chains, thereby preventing the erroneous merging of minor threads into major ones. 
We also map the nested narrative hierarchy (e.g., extradiegetic, intradiegetic, or metadiegetic levels) based on the theory of narrative levels \cite{genette1980narrative} to systematically prevent context collisions caused by embedded narratives. 
Finally, to capture the multidimensionality of narrative discourse, we build upon the three core narrative dimensions: time, setting, and point-of-view, which \citet{piper2024using} operationalized for LLMs under \citet{genette1980narrative}'s classical narratological framework. By combining these dimensions with \citet{vossen2021narratology}'s narratology-based framework for storyline extraction, we are able to comprehensively extract timelines, causelines, and critical shift points in time, space, and perspective from this structured thread data.

\paragraph{Locate Key Turning Points and Execute Segmentation}
Ultimately, the framework utilizes the established thread data to determine logical breakpoints. Turning points are treated as crucial functional shifts rather than mere plot occurrences
\cite{thompson1999storytelling}, ensuring that chunk boundaries are dictated by structural narrative transitions rather than arbitrary length constraints or linear episodic prolongations. 
Building upon this rationale, our framework segments narratives through a four-tiered location strategy. At the macro-structural level, we integrate Aristotelian plot structure theory, specifically the narrative shifts of reversal and recognition \cite{butcher1902poetics,ShenWang2010}, with turning-point identification \cite{papalampidi2019movie,hauge2017storytelling} to capture major narrative pivots: opportunity, change of plans, point of no return, major setback and reversal, climax and recognition, and aftermath and buffer. 
At the contextual level, we leverage the established multidimensional thread data to pinpoint breakpoints at transitions in time, space, or perspective, as well as shifts between parallel or nested narrative levels. At the micro-dynamic level, we align segmentation with state transitions (from equilibrium to disequilibrium to a new equilibrium) \cite{todorov1969structural} and the logic of narrative possibilities \cite{bremond1980logic,ShenWang2010}, capturing the precise moments where the fulfillment or failure of an action is resolved.
Finally, to pinpoint exact sentence-level boundaries within these general turning points, we draw inspiration from mutual information (MI) theory \cite{wagner2024automatic,wang2023m3seg} to formulate an LLM-assessable metric for semantic cohesion. The model is prompted to actively evaluate the semantic dependency between adjacent sentences, avoiding segmenting within High Semantic Cohesion Zones (e.g., spans with dense MI markers, such as continuous pronoun chains or adjacency pairs), and instead targeting Low Semantic Cohesion Zones, where semantic dependency is minimal.
Collectively, this four-tiered strategy ensures a rigorous alignment between segmentation and the underlying narrative architecture, thereby ensuring that the resulting chunks are narratively self-contained and semantically coherent.

\subsubsection{Implementation of LitSeg}

The LitSeg pipeline comprises three sequential phases: preprocessing, multi-round generation, and validation with fallback mechanisms. More implementation details are in Appendices~\ref{app:prompt-litseg} and~\ref{app:teacher-data-generation}.

\paragraph{Preprocessing}
To preserve authorial intent and semantic integrity, we first perform structure-driven splitting by isolating top-level narrative units (e.g., chapters, acts, scenes, sections, cantos, or titled episodes). This ensures that the model receives self-contained structural units while naturally accommodating the context window limits for most texts. When top-level unit metadata is unavailable, or when a narrative unit exceeds a length threshold $n$, we apply recursive word-level splitting, hierarchically segmenting the text by paragraphs and then sentences while greedily satisfying the $n$-word constraint. We set $n$ to the largest value the model's context budget allows without substantial long-context degradation. Subsequently, we assign a unique sequential index to each sentence in the candidate chunk.

\paragraph{Multi-stage Generation}
We implement a three-stage pipeline corresponding to the narratological workflow above. All stages share a unified system prompt containing the full narratological workflow, providing the model with global task awareness. The user prompt for each stage contains stage-specific formatting instructions alongside the model's outputs from preceding stages. This design implements an explicit multi-stage prompting strategy, which avoids the token overhead of unconstrained Chain-of-Thought (CoT) rationales while ensuring high-quality segmentation.
By constraining the model to output sentence indices rather than verbatim text, we enable flexible context selection while significantly reducing inference overhead. Specifically, for each segmented plot block, the model outputs a main range, represented as a closed interval between its start and end sentence indices, and an optional context list, represented as a sequence of discrete sentence indices. Sentences shared by overlapping main ranges are retained in all corresponding blocks, allowing boundary sentences to serve as shared narrative context. Unlike traditional chunking methods that rely on fixed, adjacent overlaps, this index-based representation allows the model to flexibly incorporate non-adjacent sentences from anywhere within the indexed candidate chunk that are crucial for understanding the main block. Additionally, we task the model with generating a concise subtitle for each block, leveraging its generative capabilities to summarize the core event without the burden of extensive text generation.
To constrain event granularity, we also introduce a dataset-calibrated soft length reference derived from the quantile distribution of segment lengths produced by LitSeg when the prompt contains no explicit length constraint.

\paragraph{Validation, Retry, and Fallback}
We apply dual-level validation to guarantee reliability. Syntax validation ensures that the output JSON conforms to the predefined schema. Semantic validation verifies that all main ranges and context indices are valid and that the main ranges, with overlaps permitted, collectively and exhaustively cover all source sentences, ensuring zero information loss. Invalid outputs are retried up to $N$ times. If coverage gaps persist after retries, a fallback mechanism greedily appends the missing sentences to their left-adjacent segment in source order (or the right one if unavailable). If the output remains unparsable after all retries, the system falls back to yielding the whole preprocessed chunk.

\paragraph{Reconstruction}
During text reconstruction, sentences specified by the context list, except those already included in the corresponding main range, are ordered by their original positions and prepended or appended to the main text block accordingly. The generated subtitle is then prepended to form the final context-enriched chunk.

\subsection{LitSeg-Lite: Distillation of Lightweight Chunker}
\label{subsec:litseg-lite}

With the objective of improving inference efficiency and facilitating deployment, we develop LitSeg-Lite, a compact student model distilled from LitSeg. Given the same sentence-indexed candidate chunk as LitSeg, LitSeg-Lite predicts a JSON output containing the internally summarized narratological workflow and the final index-based segmentation within a single inference pass. We train LitSeg-Lite with LoRA~\cite{DBLP:journals/corr/abs-2106-09685} on the LitSeg-annotated LiteraryQA dataset through a two-stage optimization pipeline: initial supervised fine-tuning (SFT) followed by Reinforcement Learning (RL) refinement. The SFT targets are the validated LitSeg annotations, reformatted into the single-stage output schema used by LitSeg-Lite.

To ensure that LitSeg-Lite acquires narratology-aware segmentation logic and robust instruction-following capabilities, we first perform SFT. While SFT facilitates rapid task adaptation, it often results in limited out-of-distribution generalization~\cite{chu2025sftmemorizesrlgeneralizes}. To overcome these limitations, we integrate RL to further enhance the model's behavioral robustness and generalization. Specifically, we employ GRPO with a DAPO-style objective, guided by a specialized reward function grounded in narratological theory. This function combines a rule-based format validator with a model-based reward evaluator; the latter conducts granular assessments across three narrative-centric dimensions: Event Validity, Unity of Action, and Cut Point Logic, thereby targeting common weaknesses of the student model. The model-based scores are first averaged equally across the three dimensions and then over predicted chunks to form a narratological reward, which is combined with the rule-based format reward through a weighted sum as the final optimization objective. This reward design explicitly aligns the student model with narratological criteria for structural analysis. See Appendix~\ref{app:litseg-lite-training-details} for training configurations and reward details and Appendix~\ref{app:prompt-reward} for the comprehensive scoring rubrics.

While LitSeg achieves high-fidelity results through a sophisticated three-stage prompting pipeline, such a multi-turn process introduces significant computational overhead. For practical deployment, LitSeg-Lite performs segmentation with a single model call, eliminating LitSeg's multi-stage prompting overhead while maintaining competitive performance. This offers a practical balance between narrative depth and operational efficiency. For prompts for LitSeg-Lite, see Appendix~\ref{app:prompt-litseg-lite}. For inference configurations, see Appendix~\ref{app:litseg-lite-inf-impl}.

\section{Experiments}
\label{sec:exp}

In this section, we comprehensively evaluate LitSeg and LitSeg-Lite on two narrative QA benchmarks, assessing the intrinsic quality and structural independence of their text chunks (Section~\ref{subsec:intrinsic-eval}), investigating their retrieval and generation performance (Section~\ref{subsec:exp-retrieval-generation}), validating the contribution of core design choices through ablation studies (Subsection~\ref{subsec:exp-ablation}), and demonstrating LitSeg-Lite's computational efficiency (Section~\ref{subsec:exp-eff-analysis}). Details of our RAG pipeline are provided in Appendix~\ref{app:rag-pipeline}.

\paragraph{Datasets and Baselines}
We employ two benchmarks for evaluation. LiteraryQA~\cite{bonomo-etal-2025-literaryqa} is a long-document narrative QA dataset focusing on diverse literary works (e.g., novels, narrative poetry, and drama), with all reported results evaluated on its official test split. However, as LiteraryQA lacks gold evidence-span annotations for precise retrieval evaluation, and to assess the generalization performance of LitSeg-Lite, which is trained on LitSeg annotations of the LiteraryQA training set in our experiments, we further adopt GutenQA~\cite{duarte-etal-2024-lumberchunker}, a retrieval-focused benchmark featuring ``needle-in-a-haystack'' factual QA. We compare our approaches against heuristic, embedding-based, and LLM-based segmentation baselines.
Detailed baseline configurations are provided in Appendices~\ref{app:baseline-overview} and~\ref{app:baseline-details}.

\subsection{Segment Quality}
\label{subsec:intrinsic-eval}

We evaluate the segmentation quality using both human judgment and automatic metrics to characterize the narrative coherence and self-containment of the chunks yielded by our methods. As automatic metrics are limited in capturing narratological nuances, we introduce a human evaluation protocol focusing on Content Completeness, Topical Coherence, and Boundary Placement. A panel of literary experts rates sampled chunks on a 1–5 scale, and we report per-dimension and overall averages. For automatic evaluation, we measure the boundary semantic transition by computing the cosine similarity between sentences adjacent to each predicted boundary, where lower values signify clearer narrative transitions. See Appendix~\ref{app:chunk-quality-evaluation} for human evaluation details.

\begin{table}[h]
\centering
\scriptsize
\setlength{\tabcolsep}{2.0pt}
\begin{tabular}{lcccccccc}
\toprule
\textbf{Method} & \multicolumn{4}{c}{\textbf{GutenQA}} & \multicolumn{4}{c}{\textbf{LiteraryQA}} \\
\cmidrule(lr){2-5} \cmidrule(lr){6-9}
& \textbf{Comp.} & \textbf{Coh.} & \textbf{Bnd.} & \textbf{Ovr.} & \textbf{Comp.} & \textbf{Coh.} & \textbf{Bnd.} & \textbf{Ovr.} \\
\midrule
\multicolumn{9}{l}{\emph{RAG Baselines}} \\
Token
& 3.62 & 4.32 & 2.45 & 3.46 & 3.61 & 4.19 & 2.44 & 3.43 \\
Recursive Char
& 3.65 & 4.60 & 3.61 & 3.97 & 3.65 & 4.56 & 3.55 & 3.92 \\
Perplexity
& 3.24 & 4.39 & 2.90 & 3.53 & 3.15 & 4.30 & 2.87 & 3.44 \\
\hspace{0.5em}+ Merge
& 3.72 & 4.33 & 3.09 & 3.70 & 3.74 & 4.24 & 3.20 & 3.72 \\
Margin-Sampling
& 2.23 & 4.09 & 2.89 & 3.09 & 2.09 & 4.18 & 2.81 & 3.03 \\
\hspace{0.5em}+ Merge
& 3.85 & 4.60 & 4.26 & 4.23 & 3.90 & 4.52 & 4.26 & 4.23 \\
LumberChunker
& 3.44 & 4.38 & 3.92 & 3.94 & 3.46 & 4.28 & 3.89 & 3.88 \\
\hspace{0.5em}+ Merge
& 3.97 & 4.58 & 4.34 & 4.30 & 3.99 & 4.57 & 4.39 & 4.32 \\
\midrule
\multicolumn{9}{l}{\emph{Ours}} \\
LitSeg-Lite
& \textbf{4.24} & 4.71 & \textbf{4.74} & \textbf{4.57} & \textbf{4.32} & 4.64 & \textbf{4.66} & 4.54 \\
LitSeg
& \textbf{4.24} & \textbf{4.75} & 4.70 & 4.56 & 4.30 & \textbf{4.68} & \textbf{4.66} & \textbf{4.56} \\
\bottomrule
\end{tabular}
\caption{Human segmentation quality evaluation (1--5 scale). Comp.~=~Content Completeness; Coh.~=~Topical Coherence; Bnd.~=~Boundary Placement; Ovr.~=~Overall. Best results are in \textbf{bold}.}
\label{tab:chunk_human}
\end{table}

\begin{table}[h]
\centering
\scriptsize
\begin{tabular}{lcc}
\toprule
\textbf{Method} & \textbf{GutenQA} & \textbf{LiteraryQA} \\
\midrule
\multicolumn{3}{l}{\emph{RAG Baselines}} \\
Token & 0.4536 & 0.4047 \\
Recursive Char & 0.4382 & 0.4219 \\
Perplexity & 0.4257 & 0.4064 \\
\hspace{0.5em}+ Merge & 0.4378 & 0.4179 \\
Margin-Sampling & 0.4365 & 0.4148 \\
\hspace{0.5em}+ Merge & 0.4468 & 0.4276 \\
LumberChunker & 0.4135 & 0.3805 \\
\hspace{0.5em}+ Merge & 0.4096 & 0.3761 \\
\midrule
\multicolumn{3}{l}{\emph{Ours}} \\
LitSeg-Lite & \textbf{0.4025} & \textbf{0.3699} \\
LitSeg & --- & 0.3768 \\
\bottomrule
\end{tabular}
\caption{Average semantic similarity of adjacent sentences across predicted segment boundaries. Lower values indicate better segmentation independence. Best results are in \textbf{bold}.}
\label{tab:boundary_similarity}
\end{table}

As shown in Tables~\ref{tab:chunk_human} and~\ref{tab:boundary_similarity}, LitSeg and LitSeg-Lite outperform all baselines on overall quality (e.g., 4.56/4.54 on LiteraryQA versus 4.32 for LumberChunker+Merge and 3.92 for Recursive Character; 4.56/4.57 versus 4.30 and 3.97 on GutenQA), as well as on all dimensions. The largest gap is Content Completeness, the dimension that measures whether a chunk is a self-contained narrative event, where LitSeg-Lite scores 4.32 on LiteraryQA versus 3.46 for LumberChunker, a strong LLM-based baseline. LitSeg-Lite matches the teacher on both benchmarks. LitSeg-Lite also attains the lowest boundary similarity (0.4025 on GutenQA and 0.3699 on LiteraryQA).
These empirical results demonstrate that our approach successfully identifies profound plot transitions and narrative boundaries, thereby yielding chunks that preserve narrative consistency and are intelligible on their own, and validating the effectiveness of our narrative-theory-guided segmentation framework.

\subsection{Retrieval and Generation Performance}
\label{subsec:exp-retrieval-generation}

We comprehensively assess retrieval and generation using lexical and LLM-based automatic metrics. While reliable for structured, information-centric domains with well-defined evidence boundaries, these metrics systematically penalize answers containing enriched, contextually appropriate information not included in ground truths, as they only measure surface-level literal or factual matching against ground truths~\cite{gerrits2026creativitybiasmachineevaluation}, hindering their validity on nuanced literary QA tasks. For instance, on GutenQA, our richer narrative context prompts the generator to produce highly accurate and informative, even in-depth answers. This creates a systematic discrepancy with the concise, factoid-style reference labels, which are referenced by automated metrics in isolation from the broader context, thereby underestimating our method's performance.
Similar phenomena have been observed in literary translation~\cite{zhang-etal-2025-good} and other expert knowledge tasks~\cite{10.1145/3708359.3712091}.
To obtain a reliable assessment, we therefore conduct a human pairwise evaluation on both benchmarks. Experts compare LitSeg-Lite's outputs against those of each comparison method, with both judged against the original text within a $\pm$500-character context window on GutenQA, or the entire source book on LiteraryQA. The primary criterion is factual accuracy, followed by informativeness. See Appendices~\ref{app:metric-details} and~\ref{app:metrics-impl} for metric definitions and implementation details, and Appendix~\ref{appx:pairwise-evaluation} for the full annotation guidelines and inter-annotator agreement.

Furthermore, our manual audit uncovers five categories of inherent flaws in the existing benchmarks that further undermine automated evaluation: 1) ambiguous or erroneous questions; 2) incomplete ground truth; 3) literal extractions misinterpreting narrative subtext; 4) narrow reference answers; and 5) entity and granularity mismatches. Detailed case studies are provided in Appendix~\ref{app:case-study-benchmark}.

\begin{table}[h]
\centering
\scriptsize
\setlength{\tabcolsep}{2pt}
\begin{tabular}{lccccccccc}
\toprule
\textbf{Method} 
& \multicolumn{7}{c}{\textbf{GutenQA}} 
& \multicolumn{1}{c}{\textbf{LitQA}} \\
\cmidrule(lr){2-8} \cmidrule(lr){9-9}
& \textbf{CR} & \textbf{MRR} & \textbf{H@1} & \textbf{H@2} & \textbf{H@3} & \textbf{H@5} & \textbf{H@20}
& \textbf{CR} \\
\midrule
\multicolumn{9}{l}{\emph{RAG Baselines}} \\
Token 
& 0.885 & 0.572 & 0.478 & 0.589 & 0.654 & 0.728 & 0.728
& 0.654 \\
Recursive Char 
& 0.871 & 0.614 & 0.531 & 0.631 & 0.689 & 0.748 & 0.748
& 0.664 \\
Perplexity 
& 0.861 & 0.555 & 0.463 & 0.573 & 0.640 & 0.702 & 0.702
& 0.665 \\
\hspace{0.5em}+ Merge 
& 0.889 & 0.599 & 0.514 & 0.619 & 0.679 & 0.734 & 0.734
& 0.724 \\
Margin-Sampling 
& 0.781 & 0.383 & 0.309 & 0.392 & 0.446 & 0.510 & 0.510
& 0.564 \\
\hspace{0.5em}+ Merge 
& 0.877 & 0.593 & 0.509 & 0.605 & 0.668 & 0.732 & 0.732
& 0.714 \\
LumberChunker 
& 0.885 & 0.633 & 0.552 & 0.652 & 0.708 & 0.761 & 0.761
& 0.739 \\
\hspace{0.5em}+ Merge 
& 0.893 & 0.656 & 0.574 & 0.674 & 0.736 & 0.785 & 0.785
& 0.752 \\
\midrule
\multicolumn{9}{l}{\emph{Ours}} \\
LitSeg-Lite 
& \textbf{0.938} & \textbf{0.683} & \textbf{0.600} & \textbf{0.701} & \textbf{0.759} & \textbf{0.813} & \textbf{0.813}
& \textbf{0.805} \\
LitSeg 
& --- & --- & --- & --- & --- & --- & ---
& 0.799 \\
\bottomrule
\end{tabular}
\caption{Retrieval results on GutenQA and LiteraryQA. CR~= Context Relevance; MRR~= Mean Reciprocal Rank; H@$k$~= Hit@$k$. Best results are in \textbf{bold}.}
\label{tab:retrieval_main}
\end{table}

\begin{table}[h]
\centering
\scriptsize
\setlength{\tabcolsep}{4pt}
\setlength{\tabcolsep}{2.5pt}
\begin{tabular}{lccccccc}
\toprule
\textbf{Method} & \textbf{GutenQA} & \multicolumn{6}{c}{\textbf{LiteraryQA}} \\
\cmidrule(lr){2-2} \cmidrule(lr){3-8}
& \textbf{WR} & \textbf{WR} & \textbf{EM} & \textbf{F1} & \textbf{R-L} & \textbf{Mtr} & \textbf{AA} \\
\midrule
\multicolumn{8}{l}{\emph{RAG Baselines}} \\
Token  
& 0.583 & 0.618 & 0.066 & 0.240 & 0.248 & 0.296 & 0.418 \\
Recursive Char  
& 0.552 & 0.603 & 0.078 & 0.253 & 0.259 & 0.298 & 0.415 \\
Perplexity  
& 0.603 & 0.635 & 0.072 & 0.235 & 0.245 & 0.280 & 0.373 \\
\hspace{0.5em}+ Merge  
& 0.618 & 0.628 & 0.079 & 0.253 & 0.259 & 0.302 & 0.423 \\
Margin-Sampling  
& 0.652 & 0.703 & 0.065 & 0.205 & 0.216 & 0.247 & 0.308 \\
\hspace{0.5em}+ Merge  
& 0.594 & 0.672 & 0.080 & 0.255 & 0.262 & 0.299 & 0.418 \\
LumberChunker  
& 0.584 & 0.598 & 0.076 & 0.250 & 0.255 & 0.299 & 0.435 \\
\hspace{0.5em}+ Merge  
& 0.588 & 0.540 & 0.077 & 0.252 & 0.257 & 0.303 & 0.438 \\
\midrule
\multicolumn{8}{l}{\emph{Ours}} \\
LitSeg-Lite  
& * & * & 0.079 & \textbf{0.258} & \textbf{0.263} & \textbf{0.311} & 0.448 \\
LitSeg  
& --- & 0.520 & \textbf{0.081} & 0.256 & 0.261 & 0.309 & \textbf{0.450} \\
\bottomrule
\end{tabular}
\caption{Generation results on GutenQA and LiteraryQA. EM~= Exact Match; R-L~= ROUGE-L; Mtr~= METEOR; AA~= Answer Accuracy; WR~= Pairwise Win Rate of $*$. Best results are in \textbf{bold}.}
\label{tab:literaryqa_gen}
\end{table}

Tables~\ref{tab:retrieval_main} and~\ref{tab:literaryqa_gen} report the retrieval and generation results on both benchmarks.
Our proposed methods consistently achieve competitive performance across retrieval and generation metrics.
Notably, significant gains are observed in Context Relevance, where LitSeg-Lite outperforms the best baselines by substantial margins (+0.045 on GutenQA; +0.053 on LiteraryQA), demonstrating that narratological boundaries effectively preserve plot-complete passages.
This segmentation advantage is further reflected in retrieval performance, evidenced by the highest MRR (0.683) and substantial improvements in Hit@$k$ metrics (e.g., 0.813 on H@5 for GutenQA), indicating that our method not only retrieves the correct chunks but also ranks them significantly higher.
Consequently, this precise retrieval and prioritization of relevant context empowers the generator to utilize necessary information, confirmed by LitSeg achieving the highest Answer Accuracy on LiteraryQA (0.450), surpassing the best baseline (0.438). Pairwise human evaluation further shows that LitSeg-Lite is preferred over all baselines on both benchmarks (e.g., a 59.8\% win rate against LumberChunker on LiteraryQA and a 65.2\% win rate against Margin-Sampling on GutenQA).
Furthermore, the distillation process proves highly effective, as LitSeg-Lite not only matches but slightly exceeds LitSeg's Context Relevance (0.805 vs. 0.799) and achieves comparable generation quality (Answer Accuracy of 0.448 vs. 0.450; pairwise win rate of 0.520 on LiteraryQA). This confirms that LitSeg-Lite successfully internalizes the narrative-aware segmentation logic of LitSeg in a single, efficient pass using a compact model.

To qualitatively illustrate how our framework preserves narrative integrity and enhances downstream QA performance, we provide case studies and detailed textual analyses in Appendix~\ref{app:case-study-chunking}. See Appendix~\ref{app:gutenqa_automated} for more quantitative metrics.

\subsection{Ablation Studies}
\label{subsec:exp-ablation}

To validate the contributions of distilled knowledge and narratological theory, we ablate LitSeg-Lite into six variants: \textit{w/o GRPO} (retains SFT but removes the RL stage), \textit{w/o FT} (retains the theory prompt but removes fine-tuning), \textit{+w/o Theo.S.1/2/3} (each ablates one specific theory-driven stage beyond \textit{w/o FT}), and \textit{+w/o Theory} (replaces the entire narratological pipeline with a generic prompt beyond \textit{w/o FT}). See Appendix~\ref{app:prompt-ablation} for detailed prompt templates.

\begin{table}[h]
\centering
\scriptsize
\setlength{\tabcolsep}{2.0pt}
\begin{tabular}{lcccccccc}
\toprule
\textbf{Method} & \multicolumn{4}{c}{\textbf{GutenQA}} & \multicolumn{4}{c}{\textbf{LiteraryQA}} \\
\cmidrule(lr){2-5} \cmidrule(lr){6-9}
& \textbf{Comp.} & \textbf{Coh.} & \textbf{Bnd.} & \textbf{Ovr.} & \textbf{Comp.} & \textbf{Coh.} & \textbf{Bnd.} & \textbf{Ovr.} \\
\midrule
LitSeg-Lite
& \textbf{4.24} & \textbf{4.71} & \textbf{4.74} & \textbf{4.57} & \textbf{4.32} & 4.64 & \textbf{4.66} & \textbf{4.54} \\
\hspace{0.5em}w/o GRPO
& 4.23 & \textbf{4.71} & 4.72 & 4.55 & 4.28 & \textbf{4.65} & 4.61 & 4.51 \\
\hspace{0.5em}w/o FT
& 4.18 & 4.53 & 4.39 & 4.37 & 4.19 & 4.44 & 4.28 & 4.30 \\
\hspace{0.5em}+w/o Theo.S.1
& 4.12 & 4.48 & 4.33 & 4.32 & 4.13 & 4.40 & 4.25 & 4.27 \\
\hspace{0.5em}+w/o Theo.S.2
& 4.08 & 4.46 & 4.37 & 4.31 & 4.13 & 4.40 & 4.28 & 4.26 \\
\hspace{0.5em}+w/o Theo.S.3
& 3.93 & 4.52 & 4.24 & 4.23 & 4.07 & 4.46 & 4.25 & 4.26 \\
\hspace{0.5em}+w/o Theory
& 3.83 & 4.65 & 4.28 & 4.25 & 3.93 & 4.56 & 4.27 & 4.25 \\
\bottomrule
\end{tabular}
\caption{Ablation study on human evaluation of segmentation quality. Best results are in \textbf{bold}.}
\label{tab:chunk_human_ablation}
\end{table}

\begin{table}[h]
\centering
\scriptsize
\begin{tabular}{lcc}
\toprule
\textbf{Method} & \textbf{GutenQA} & \textbf{LiteraryQA} \\
\midrule
LitSeg-Lite & \textbf{0.4025} & \textbf{0.3699} \\
\hspace{0.5em}w/o GRPO & 0.4048 & 0.3711 \\
\hspace{0.5em}w/o FT & 0.4350 & 0.3989 \\
\hspace{0.5em}+w/o Theo.S.1 & 0.4349 & 0.3941 \\
\hspace{0.5em}+w/o Theo.S.2 & 0.4385 & 0.4012 \\
\hspace{0.5em}+w/o Theo.S.3 & 0.4403 & 0.4031 \\
\hspace{0.5em}+w/o Theory & 0.4324 & 0.3991 \\
\bottomrule
\end{tabular}
\caption{Ablation study on average boundary semantic similarity. Best results are in \textbf{bold}.}
\label{tab:segmentation_ablation}
\end{table}

\begin{table}[h]
\centering
\scriptsize
\setlength{\tabcolsep}{2pt}
\begin{tabular}{lccccccccc}
\toprule
\textbf{Method} 
& \multicolumn{7}{c}{\textbf{GutenQA}} 
& \multicolumn{1}{c}{\textbf{LitQA}} \\
\cmidrule(lr){2-8} \cmidrule(lr){9-9}
& \textbf{CR} & \textbf{MRR} & \textbf{H@1} & \textbf{H@2} & \textbf{H@3} & \textbf{H@5} & \textbf{H@20}
& \textbf{CR} \\
\midrule
LitSeg-Lite 
& \textbf{0.938} & \textbf{0.683} & \textbf{0.600} & \textbf{0.701} & \textbf{0.759} & \textbf{0.813} & \textbf{0.813}
& \textbf{0.805} \\
\hspace{0.5em}w/o GRPO 
& 0.937 & 0.675 & 0.588 & 0.698 & 0.753 & 0.812 & 0.812
& 0.795 \\
\hspace{0.5em}w/o FT 
& 0.927 & 0.646 & 0.550 & 0.670 & 0.737 & 0.797 & 0.796
& 0.787 \\
\hspace{0.5em}+w/o Theo.S.1 
& 0.927 & 0.634 & 0.532 & 0.656 & 0.728 & 0.794 & 0.794
& 0.791 \\
\hspace{0.5em}+w/o Theo.S.2 
& 0.922 & 0.628 & 0.527 & 0.647 & 0.719 & 0.792 & 0.792
& 0.781 \\
\hspace{0.5em}+w/o Theo.S.3 
& 0.925 & 0.625 & 0.531 & 0.648 & 0.709 & 0.774 & 0.774
& 0.771 \\
\hspace{0.5em}+w/o Theory 
& 0.936 & 0.632 & 0.536 & 0.657 & 0.720 & 0.783 & 0.783
& 0.764 \\
\bottomrule
\end{tabular}
\caption{Ablation study on GutenQA and LiteraryQA retrieval. Best results are in \textbf{bold}.}
\label{tab:retrieval_ablation}
\end{table}

\begin{table}[h]
\centering
\scriptsize
\setlength{\tabcolsep}{4pt}
\setlength{\tabcolsep}{2.5pt}
\begin{tabular}{lccccccc}
\toprule
\textbf{Method} & \textbf{GutenQA} & \multicolumn{6}{c}{\textbf{LiteraryQA}} \\
\cmidrule(lr){2-2} \cmidrule(lr){3-8}
& \textbf{WR} & \textbf{WR} & \textbf{EM} & \textbf{F1} & \textbf{R-L} & \textbf{Mtr} & \textbf{AA} \\
\midrule
LitSeg-Lite  
& * & * & \textbf{0.079} & \textbf{0.258} & \textbf{0.263} & \textbf{0.311} & \textbf{0.448} \\
\hspace{0.5em}w/o GRPO  
& 0.535 & 0.547 & 0.079 & 0.252 & 0.257 & 0.307 & 0.446 \\
\hspace{0.5em}w/o FT  
& 0.593 & 0.548 & 0.077 & 0.253 & 0.256 & 0.306 & 0.446 \\
\hspace{0.5em}+w/o Theo.S.1  
& 0.553 & 0.555 & 0.074 & 0.248 & 0.252 & 0.298 & 0.441 \\
\hspace{0.5em}+w/o Theo.S.2  
& 0.564 & 0.600 & 0.076 & 0.250 & 0.254 & 0.300 & 0.441 \\
\hspace{0.5em}+w/o Theo.S.3  
& 0.548 & 0.600 & 0.078 & 0.255 & 0.259 & 0.308 & 0.445 \\
\hspace{0.5em}+w/o Theory  
& 0.584 & 0.613 & 0.073 & 0.250 & 0.255 & 0.301 & 0.438 \\
\bottomrule
\end{tabular}
\caption{Ablation study on GutenQA and LiteraryQA generation. Best results are in \textbf{bold}.}
\label{tab:literaryqa_gen_ablation}
\end{table}

As shown in Tables~\ref{tab:chunk_human_ablation},~\ref{tab:segmentation_ablation},~\ref{tab:retrieval_ablation} and~\ref{tab:literaryqa_gen_ablation}, both distillation and the multi-step narratological pipeline contribute to the overall performance of LitSeg-Lite. Removing fine-tuning stages or skipping any individual theory-driven stage consistently degrades performance.
For segmentation quality, expert overall scores on LiteraryQA drop from 4.54 to 4.51 (\textit{w/o GRPO}) and 4.30 (\textit{w/o FT}); ablating any single theory stage yields 4.26--4.27, while removing the entire pipeline yields 4.25. Boundary similarity similarly increases with the removal of either fine-tuning or narratological prompting from 0.3699 (\textit{LitSeg-Lite}) to 0.3711 (\textit{w/o GRPO}) and 0.3989 (\textit{w/o FT}), and further to 0.3991 (\textit{+w/o Theory}), confirming that our framework is essential for regularizing the model to produce narratologically self-contained and coherent text chunks.
Degradation is also observed in retrieval performance. On GutenQA, MRR falls from 0.683 (\textit{LitSeg-Lite}) to 0.675 (\textit{w/o GRPO}) and 0.646 (\textit{w/o FT}), then to 0.634, 0.628 and 0.625 (\textit{+w/o Theo.S.1--3}) and 0.632 (\textit{+w/o Theory}). A similar pattern emerges on LiteraryQA, where Context Relevance degrades from 0.805 to 0.795 (\textit{w/o GRPO}), 0.787 (\textit{w/o FT}), and further to 0.764 (\textit{+w/o Theory}), indicating the importance of fine-tuning and theory guidance for retrieval quality.
For downstream generation, Answer Accuracy on LiteraryQA decreases from 0.448 to 0.438 (\textit{+w/o Theory}), while pairwise evaluation prefers the full model on both benchmarks (e.g., a 53.5\% win rate against \textit{w/o GRPO} and 58.4\% against \textit{+w/o Theory} on GutenQA; 54.7\% against \textit{w/o GRPO} and 61.3\% against \textit{+w/o Theory} on LiteraryQA).
Ablating one component sometimes even incurs degraded performance compared to \textit{+w/o Theory}, confirming that the holistic integration of both distillation and theory-guided pipelines contributes to chunk quality that facilitates document retrieval and generation.

\subsection{Efficiency Analysis}
\label{subsec:exp-eff-analysis}

We time end-to-end chunking of the LiteraryQA test set for LitSeg, LitSeg-Lite, and LumberChunker (the overall strongest baseline) on a single RTX 6000 Ada GPU with vLLM~\cite{kwon2023efficient}, with LitSeg-Lite and LumberChunker using the same inference configuration. LitSeg-Lite completes in 9,442\,s, 2.69$\times$ faster than LitSeg (25,362\,s; 37\% of the teacher's compute) and 1.43$\times$ faster than LumberChunker (13,514\,s), while matching the teacher's quality and outperforming LumberChunker on segmentation quality and downstream QA. These results confirm that distillation yields a lightweight chunker with substantially lower inference cost.

\section{Conclusion}
\label{sec:conclusion}
In this work, we address the critical challenge of document segmentation for literary RAG systems, where existing methods overlook high-order narrative structures, leading to fragmented contexts and degraded QA performance. We introduce LitSeg, a narrative-theory-guided framework employing multi-stage prompting to explicitly extract narrative structures, complemented by a novel sentence-index-based output schema that enables flexible context incorporation and reduces generation overhead. To mitigate the computational cost of this multi-stage pipeline, we further present LitSeg-Lite, a lightweight single-pass chunker distilled via a two-stage SFT and RL strategy. Extensive experiments on GutenQA and LiteraryQA demonstrate that our proposed methods substantially outperform baselines, yielding high-quality text chunks that are narratologically coherent and self-contained, facilitating gains in context relevance, retrieval accuracy, and downstream answer accuracy. Ablation studies confirm the substantial contributions of both narratological guidance and knowledge distillation, and efficiency analysis confirms the practical advantages of LitSeg-Lite. Together, LitSeg and LitSeg-Lite offer a narrative-aware segmentation framework that enhances RAG deployment, effectively bridging the gap between literary theory and practical applications.

\section*{Limitations}

Although our evaluation focuses on literary RAG, the narratologically self-contained chunks produced by LitSeg could potentially benefit other literary NLP tasks, such as sentiment and emotion analysis~\cite{kim2021survey} and narrative-arc computation~\cite{doi:10.1126/sciadv.aba2196}.

Our evaluation is also constrained by limitations in existing literary benchmarks. Quantitative metrics may not fully capture the extent of our method's improvements in certain cases. To provide a more comprehensive view, we present a detailed qualitative analysis in Appendix~\ref{app:case-study-benchmark}.

\bibliography{custom}

@article{Zhao2026RAG,
  author = {Zhao, Penghao and Zhang, Hailin and Yu, Qinhan and Wang, Zhengren and Geng, Yunteng and Fu, Fangcheng and Yang, Ling and Zhang, Wentao and Jiang, Jie and Cui, Bin},
  title = {Retrieval-Augmented Generation for AI-Generated Content: A Survey},
  journal = {Data Science and Engineering},
  volume = {11},
  number = {1},
  pages = {1--29},
  year = {2026},
  issn = {2364-1541},
  url = {https://doi.org/10.1007/s41019-025-00335-5},
  doi = {10.1007/s41019-025-00335-5}
}

@inproceedings{wang-etal-2025-document,
    title = "Document Segmentation Matters for Retrieval-Augmented Generation",
    author = "Wang, Zhitong  and
      Gao, Cheng  and
      Xiao, Chaojun  and
      Huang, Yufei  and
      Si, Shuzheng  and
      Luo, Kangyang  and
      Bai, Yuzhuo  and
      Li, Wenhao  and
      Duan, Tangjian  and
      Lv, Chuancheng  and
      Lu, Guoshan  and
      Chen, Gang  and
      Qi, Fanchao  and
      Sun, Maosong",
    editor = "Che, Wanxiang  and
      Nabende, Joyce  and
      Shutova, Ekaterina  and
      Pilehvar, Mohammad Taher",
    booktitle = "Findings of the Association for Computational Linguistics: ACL 2025",
    month = jul,
    year = "2025",
    address = "Vienna, Austria",
    publisher = "Association for Computational Linguistics",
    url = "https://aclanthology.org/2025.findings-acl.422/",
    doi = "10.18653/v1/2025.findings-acl.422",
    pages = "8063--8075",
    ISBN = "979-8-89176-256-5"
}

@INPROCEEDINGS{11300411,
  author={Ma, Yusong and Nie, Hongxuan and Chen, Chao and Zhang, Jiujie and Jiang, Jiali and Wang, Bisheng and Xia, Yuqin},
  booktitle={2025 International Conference on Trustworthy Big Data and Artificial Intelligence (ICTBAI)}, 
  title={A Survey of Retrieval-Augmented Generation (RAG) for Large Language Models}, 
  year={2025},
  volume={},
  number={},
  pages={7-13},
  doi={10.1109/ICTBAI68361.2025.00008}}

@InProceedings{10.1007/978-3-032-15404-0_30,
author="Rajanidi, Saikrishna
and Anbazhagan, M.
and Ramya, G. R.",
editor="Nanda, Satyasai Jagannath
and Yadav, Rajendra Prasad
and Prasad, Mukesh
and Saraswat, Mukesh",
title="RAG in Specialized Domains: A Survey of QA Chatbots",
booktitle="Data Science and Applications",
year="2026",
publisher="Springer Nature Switzerland",
address="Cham",
pages="352--369",
isbn="978-3-032-15404-0"
}

@misc{singh2025agenticretrievalaugmentedgenerationsurvey,
      title={Agentic Retrieval-Augmented Generation: A Survey on Agentic RAG}, 
      author={Aditi Singh and Abul Ehtesham and Saket Kumar and Tala Talaei Khoei},
      year={2025},
      eprint={2501.09136},
      archivePrefix={arXiv},
      primaryClass={cs.AI},
      url={https://arxiv.org/abs/2501.09136}, 
}

@inproceedings{gao-etal-2023-precise,
    title = "Precise Zero-Shot Dense Retrieval without Relevance Labels",
    author = "Gao, Luyu  and
      Ma, Xueguang  and
      Lin, Jimmy  and
      Callan, Jamie",
    editor = "Rogers, Anna  and
      Boyd-Graber, Jordan  and
      Okazaki, Naoaki",
    booktitle = "Proceedings of the 61st Annual Meeting of the Association for Computational Linguistics (Volume 1: Long Papers)",
    month = jul,
    year = "2023",
    address = "Toronto, Canada",
    publisher = "Association for Computational Linguistics",
    url = "https://aclanthology.org/2023.acl-long.99/",
    doi = "10.18653/v1/2023.acl-long.99",
    pages = "1762--1777"
}

@inproceedings{huang-etal-2025-retrieval,
    title = "Retrieval-Augmented Generation with Hierarchical Knowledge",
    author = "Huang, Haoyu  and
      Huang, Yongfeng  and
      Junjie, Yang  and
      Pan, Zhenyu  and
      Chen, Yongqiang  and
      Ma, Kaili  and
      Chen, Hongzhi  and
      Cheng, James",
    editor = "Christodoulopoulos, Christos  and
      Chakraborty, Tanmoy  and
      Rose, Carolyn  and
      Peng, Violet",
    booktitle = "Findings of the Association for Computational Linguistics: EMNLP 2025",
    month = nov,
    year = "2025",
    address = "Suzhou, China",
    publisher = "Association for Computational Linguistics",
    url = "https://aclanthology.org/2025.findings-emnlp.321/",
    doi = "10.18653/v1/2025.findings-emnlp.321",
    pages = "6044--6060",
    ISBN = "979-8-89176-335-7"
}

@software{Chase_LangChain_2022,
author = {Chase, Harrison},
month = oct,
title = {{LangChain}},
url = {https://github.com/langchain-ai/langchain},
year = {2022}
}

@inproceedings{10.1145/3726302.3730078,
author = {Tu, Yiteng and Su, Weihang and Zhou, Yujia and Liu, Yiqun and Ai, Qingyao},
title = {Robust Fine-tuning for Retrieval Augmented Generation against Retrieval Defects},
year = {2025},
isbn = {9798400715921},
publisher = {Association for Computing Machinery},
address = {New York, NY, USA},
url = {https://doi.org/10.1145/3726302.3730078},
doi = {10.1145/3726302.3730078},
booktitle = {Proceedings of the 48th International ACM SIGIR Conference on Research and Development in Information Retrieval},
pages = {1272–1282},
numpages = {11},
location = {Padua, Italy},
series = {SIGIR '25}
}

@misc{schulman2017proximalpolicyoptimizationalgorithms,
      title={Proximal Policy Optimization Algorithms}, 
      author={John Schulman and Filip Wolski and Prafulla Dhariwal and Alec Radford and Oleg Klimov},
      year={2017},
      eprint={1707.06347},
      archivePrefix={arXiv},
      primaryClass={cs.LG},
      url={https://arxiv.org/abs/1707.06347}, 
}

@misc{yu2025dapoopensourcellmreinforcement,
      title={DAPO: An Open-Source LLM Reinforcement Learning System at Scale}, 
      author={Qiying Yu and Zheng Zhang and Ruofei Zhu and Yufeng Yuan and Xiaochen Zuo and Yu Yue and Weinan Dai and Tiantian Fan and Gaohong Liu and Lingjun Liu and Xin Liu and Haibin Lin and Zhiqi Lin and Bole Ma and Guangming Sheng and Yuxuan Tong and Chi Zhang and Mofan Zhang and Wang Zhang and Hang Zhu and Jinhua Zhu and Jiaze Chen and Jiangjie Chen and Chengyi Wang and Hongli Yu and Yuxuan Song and Xiangpeng Wei and Hao Zhou and Jingjing Liu and Wei-Ying Ma and Ya-Qin Zhang and Lin Yan and Mu Qiao and Yonghui Wu and Mingxuan Wang},
      year={2025},
      eprint={2503.14476},
      archivePrefix={arXiv},
      primaryClass={cs.LG},
      url={https://arxiv.org/abs/2503.14476}, 
}

@inproceedings{zhang-etal-2025-good,
    title = "How Good Are {LLM}s for Literary Translation, Really? Literary Translation Evaluation with Humans and {LLM}s",
    author = "Zhang, Ran  and
      Zhao, Wei  and
      Eger, Steffen",
    editor = "Chiruzzo, Luis  and
      Ritter, Alan  and
      Wang, Lu",
    booktitle = "Proceedings of the 2025 Conference of the Nations of the Americas Chapter of the Association for Computational Linguistics: Human Language Technologies (Volume 1: Long Papers)",
    month = apr,
    year = "2025",
    address = "Albuquerque, New Mexico",
    publisher = "Association for Computational Linguistics",
    url = "https://aclanthology.org/2025.naacl-long.548/",
    doi = "10.18653/v1/2025.naacl-long.548",
    pages = "10961--10988",
    ISBN = "979-8-89176-189-6"
}

@inproceedings{10.1145/3708359.3712091,
author = {Szymanski, Annalisa and Ziems, Noah and Eicher-Miller, Heather A. and Li, Toby Jia-Jun and Jiang, Meng and Metoyer, Ronald A.},
title = {Limitations of the LLM-as-a-Judge Approach for Evaluating LLM Outputs in Expert Knowledge Tasks},
year = {2025},
isbn = {9798400713064},
publisher = {Association for Computing Machinery},
address = {New York, NY, USA},
url = {https://doi.org/10.1145/3708359.3712091},
doi = {10.1145/3708359.3712091},
booktitle = {Proceedings of the 30th International Conference on Intelligent User Interfaces},
pages = {952–966},
numpages = {15},
location = {
},
series = {IUI '25}
}

@misc{gerrits2026creativitybiasmachineevaluation,
      title={Creativity Bias: How Machine Evaluation Struggles with Creativity in Literary Translations}, 
      author={Kyo Gerrits and Rik van Noord and Ana Guerberof Arenas},
      year={2026},
      eprint={2605.13596},
      archivePrefix={arXiv},
      primaryClass={cs.CL},
      url={https://arxiv.org/abs/2605.13596}, 
}

@article{DBLP:journals/corr/abs-2106-09685,
  author       = {Edward J. Hu and
                  Yelong Shen and
                  Phillip Wallis and
                  Zeyuan Allen{-}Zhu and
                  Yuanzhi Li and
                  Shean Wang and
                  Weizhu Chen},
  title        = {LoRA: Low-Rank Adaptation of Large Language Models},
  journal      = {CoRR},
  volume       = {abs/2106.09685},
  year         = {2021},
  url          = {https://arxiv.org/abs/2106.09685},
  eprinttype   = {arXiv},
  eprint       = {2106.09685},
  bibsource    = {dblp computer science bibliography, https://dblp.org}
}

@article{10.1145/3777411,
author = {Zhang, Shengyu and Dong, Linfeng and Li, Xiaoya and Zhang, Sen and Sun, Xiaofei and Wang, Shuhe and Li, Jiwei and Hu, Runyi and Zhang, Tianwei and Wang, Guoyin and Wu, Fei},
title = {Instruction Tuning for Large Language Models: A Survey},
year = {2026},
issue_date = {May 2026},
publisher = {Association for Computing Machinery},
address = {New York, NY, USA},
volume = {58},
number = {7},
issn = {0360-0300},
url = {https://doi.org/10.1145/3777411},
doi = {10.1145/3777411},
journal = {ACM Comput. Surv.},
month = jan,
articleno = {169},
numpages = {36}
}

@misc{chu2025sftmemorizesrlgeneralizes,
      title={SFT Memorizes, RL Generalizes: A Comparative Study of Foundation Model Post-training}, 
      author={Tianzhe Chu and Yuexiang Zhai and Jihan Yang and Shengbang Tong and Saining Xie and Dale Schuurmans and Quoc V. Le and Sergey Levine and Yi Ma},
      year={2025},
      eprint={2501.17161},
      archivePrefix={arXiv},
      primaryClass={cs.AI},
      url={https://arxiv.org/abs/2501.17161}, 
}

@misc{shao2024deepseekmathpushinglimitsmathematical,
      title={DeepSeekMath: Pushing the Limits of Mathematical Reasoning in Open Language Models}, 
      author={Zhihong Shao and Peiyi Wang and Qihao Zhu and Runxin Xu and Junxiao Song and Xiao Bi and Haowei Zhang and Mingchuan Zhang and Y. K. Li and Y. Wu and Daya Guo},
      year={2024},
      eprint={2402.03300},
      archivePrefix={arXiv},
      primaryClass={cs.CL},
      url={https://arxiv.org/abs/2402.03300}, 
}

@inproceedings{bonomo-etal-2025-literaryqa,
    title = "{L}iterary{QA}: Towards Effective Evaluation of Long-document Narrative {QA}",
    author = "Bonomo, Tommaso  and
      Gioffr{\'e}, Luca  and
      Navigli, Roberto",
    editor = "Christodoulopoulos, Christos  and
      Chakraborty, Tanmoy  and
      Rose, Carolyn  and
      Peng, Violet",
    booktitle = "Proceedings of the 2025 Conference on Empirical Methods in Natural Language Processing",
    month = nov,
    year = "2025",
    address = "Suzhou, China",
    publisher = "Association for Computational Linguistics",
    url = "https://aclanthology.org/2025.emnlp-main.1729/",
    doi = "10.18653/v1/2025.emnlp-main.1729",
    pages = "34086--34107",
    ISBN = "979-8-89176-332-6"
}

@inproceedings{lin-2004-rouge,
    title = "{ROUGE}: A Package for Automatic Evaluation of Summaries",
    author = "Lin, Chin-Yew",
    booktitle = "Text Summarization Branches Out",
    month = jul,
    year = "2004",
    address = "Barcelona, Spain",
    publisher = "Association for Computational Linguistics",
    url = "https://aclanthology.org/W04-1013/",
    pages = "74--81"
}

@inproceedings{banerjee-lavie-2005-meteor,
    title = "{METEOR}: An Automatic Metric for {MT} Evaluation with Improved Correlation with Human Judgments",
    author = "Banerjee, Satanjeev  and
      Lavie, Alon",
    editor = "Goldstein, Jade  and
      Lavie, Alon  and
      Lin, Chin-Yew  and
      Voss, Clare",
    booktitle = "Proceedings of the {ACL} Workshop on Intrinsic and Extrinsic Evaluation Measures for Machine Translation and/or Summarization",
    month = jun,
    year = "2005",
    address = "Ann Arbor, Michigan",
    publisher = "Association for Computational Linguistics",
    url = "https://aclanthology.org/W05-0909/",
    pages = "65--72"
}

@misc{zhao2025metachunkinglearningtextsegmentation,
      title={Meta-Chunking: Learning Text Segmentation and Semantic Completion via Logical Perception}, 
      author={Jihao Zhao and Zhiyuan Ji and Yuchen Feng and Pengnian Qi and Simin Niu and Bo Tang and Feiyu Xiong and Zhiyu Li},
      year={2025},
      eprint={2410.12788},
      archivePrefix={arXiv},
      primaryClass={cs.CL},
      url={https://arxiv.org/abs/2410.12788}, 
}

@inproceedings{duarte-etal-2024-lumberchunker,
    title = "{L}umber{C}hunker: Long-Form Narrative Document Segmentation",
    author = "Duarte, Andr{\'e} V.  and
      Marques, Jo{\~a}o DS  and
      Gra{\c{c}}a, Miguel  and
      Freire, Miguel  and
      Li, Lei  and
      Oliveira, Arlindo L.",
    editor = "Al-Onaizan, Yaser  and
      Bansal, Mohit  and
      Chen, Yun-Nung",
    booktitle = "Findings of the Association for Computational Linguistics: EMNLP 2024",
    month = nov,
    year = "2024",
    address = "Miami, Florida, USA",
    publisher = "Association for Computational Linguistics",
    url = "https://aclanthology.org/2024.findings-emnlp.377/",
    doi = "10.18653/v1/2024.findings-emnlp.377",
    pages = "6473--6486"
}

@inproceedings{bird-loper-2004-nltk,
    title = "{NLTK}: The Natural Language Toolkit",
    author = "Bird, Steven  and
      Loper, Edward",
    booktitle = "Proceedings of the {ACL} Interactive Poster and Demonstration Sessions",
    month = jul,
    year = "2004",
    address = "Barcelona, Spain",
    publisher = "Association for Computational Linguistics",
    url = "https://aclanthology.org/P04-3031/",
    pages = "214--217"
}

@book{ShenWang2010,
  author = {Shen, Dan and Wang, Liya},
  title = {Western Narratology: Classical and Post-Classical},
  year = {2010},
  publisher = {Peking University Press},
  note = {In Chinese.}
}

@article{gius2022towards,
  title={Towards an event based plot model. a computational narratology approach},
  author={Gius, Evelyn and Vauth, Michael},
  journal={Journal of Computational Literary Studies},
  volume={1},
  number={1},
  year={2022},
  publisher={Universit{\"a}ts-und Landesbibliothek Darmstadt}
}

@inproceedings{piper2021narrative,
  title={Narrative theory for computational narrative understanding},
  author={Piper, Andrew and So, Richard Jean and Bamman, David},
  booktitle={Proceedings of the 2021 conference on empirical methods in natural language processing},
  pages={298--311},
  year={2021}
}

@book{genette1980narrative,
  title={Narrative discourse: An essay in method},
  author={Genette, G{\'e}rard},
  volume={3},
  year={1980},
  publisher={Cornell University Press}
}

@article{vossen2021narratology,
  title={A narratology-based framework for storyline extraction},
  author={Vossen, Piek and Caselli, Tommaso and Segers, Roxane},
  journal={Computational Analysis of Storylines: Making Sense of Events},
  volume={125},
  pages={125--140},
  year={2021},
  publisher={Cambridge University Press Cambridge}
}

@inproceedings{piper2024using,
  title={Using large language models for understanding narrative discourse},
  author={Piper, Andrew and Bagga, Sunyam},
  booktitle={Proceedings of the 6th Workshop on Narrative Understanding},
  pages={37--46},
  year={2024}
}

@book{hauge2017storytelling,
  title={Storytelling Made Easy: Persuade and Transform Your Audiences, Buyers, And Clients-Simply, Quickly, and Profitably},
  author={Hauge, Michael},
  year={2017},
  publisher={BookBaby}
}

@inproceedings{papalampidi2019movie,
  title={Movie plot analysis via turning point identification},
  author={Papalampidi, Pinelopi and Keller, Frank and Lapata, Mirella},
  booktitle={Proceedings of the 2019 Conference on Empirical Methods in Natural Language Processing and the 9th International Joint Conference on Natural Language Processing (EMNLP-IJCNLP)},
  pages={1707--1717},
  year={2019}
}

@book{thompson1999storytelling,
  title={Storytelling in the new Hollywood: Understanding classical narrative technique},
  author={Thompson, Kristin},
  year={1999},
  publisher={Harvard University Press}
}

@inproceedings{wang2023m3seg,
  title={M3Seg: A Maximum-Minimum Mutual Information Paradigm for Unsupervised Topic Segmentation in ASR Transcripts},
  author={Wang, Ke and Zhao, Xiutian and Li, Yanghui and Peng, Wei},
  booktitle={Proceedings of the 2023 Conference on Empirical Methods in Natural Language Processing},
  pages={7928--7934},
  year={2023}
}

@article{wagner2024automatic,
  title={Automatic Topic-Guided Segmentation of Holocaust Survivor Testimonies},
  author={Wagner, Eitan and Keydar, Renana and Pinchevski, Amit and Abend, Omri},
  journal={Journal of Computational Literary Studies},
  volume={2},
  number={1},
  year={2024},
  publisher={Universit{\"a}ts-und Landesbibliothek Darmstadt}
}

@inbook{Margolin+2014+777+786,
url = {https://doi.org/10.1515/9783110316469.777},
title = {Simultaneity in Narrative},
booktitle = {Handbook of Narratology},
author = {Uri Margolin},
editor = {Peter Hühn and Jan Christoph Meister and John Pier and Wolf Schmid},
publisher = {De Gruyter},
address = {Berlin, München, Boston},
pages = {777--786},
doi = {doi:10.1515/9783110316469.777},
isbn = {9783110316469},
year = {2014},
lastchecked = {2026-05-13}
}

@book{scholes2006nature,
  title={The nature of narrative: Revised and expanded},
  author={Scholes, Robert and Phelan, James and Kellogg, Robert Leland},
  year={2006},
  publisher={OUP USA}
}

@book{butcher1902poetics,
  title={The poetics of Aristotle},
  author={Butcher, Samuel Henry and others},
  year={1902},
  publisher={Macmillan}
}

@book{forster1927aspects,
  title={Aspects of the Novel},
  author={Forster, Edward Morgan},
  year={1927},
  publisher={Harcourt, Brace}
}

@article{bremond1980logic,
  title={The logic of narrative possibilities},
  author={Bremond, Claude and Cancalon, Elaine D},
  journal={New Literary History},
  volume={11},
  number={3},
  pages={387--411},
  year={1980},
  publisher={JSTOR}
}

@inproceedings{todorov1969structural,
  title={Structural analysis of narrative},
  author={Todorov, Tzvetan and Weinstein, Arnold},
  booktitle={NOVEL: A forum on fiction},
  volume={3},
  number={1},
  pages={70--76},
  year={1969},
  organization={JSTOR}
}

@book{liveley2019narratology,
  title={Narratology},
  author={Liveley, Genevieve},
  year={2019},
  publisher={Oxford University Press}
}

@article{visser2025event,
  title={Event Detection between Literary Studies and NLP: A Survey, a Narratological Reflection, and a Case Study},
  author={Visser Solissa, Noa and Cranenburgh, Andreas van and Pianzola, Federico},
  year={2025}
}

@inproceedings{kwon2023efficient,
  title={Efficient Memory Management for Large Language Model Serving with PagedAttention},
  author={Woosuk Kwon and Zhuohan Li and Siyuan Zhuang and Ying Sheng and Lianmin Zheng and Cody Hao Yu and Joseph E. Gonzalez and Hao Zhang and Ion Stoica},
  booktitle={Proceedings of the ACM SIGOPS 29th Symposium on Operating Systems Principles},
  year={2023}
}

@article{kim2021survey,
  title     = {A Survey on Sentiment and Emotion Analysis for Computational Literary Studies},
  author    = {Kim, Evgeny and Klinger, Roman},
  journal   = {Zeitschrift für digitale Geisteswissenschaften},
  year      = {2021},
  month     = jul,
  note      = {Version 2.0},
  publisher = {Wolfenbüttel},
  doi       = {10.17175/2019_008_v2},
  url       = {https://doi.org/10.17175/2019_008_v2}
}

@article{
doi:10.1126/sciadv.aba2196,
author = {Ryan L. Boyd  and Kate G. Blackburn  and James W. Pennebaker },
title = {The narrative arc: Revealing core narrative structures through text analysis},
journal = {Science Advances},
volume = {6},
number = {32},
pages = {eaba2196},
year = {2020},
doi = {10.1126/sciadv.aba2196},
URL = {https://www.science.org/doi/abs/10.1126/sciadv.aba2196},
eprint = {https://www.science.org/doi/pdf/10.1126/sciadv.aba2196}}

\appendix

\section{Details of Expert Chunk Quality Evaluation}
\label{app:chunk-quality-evaluation}

\subsection{Expert Information and Annotation Procedure}
We recruit two literary experts to rate segmentation quality. For each chunking strategy, $n=100$ chunks are sampled on each of GutenQA and LiteraryQA. Experts score every chunk independently on a 1--5 scale (higher is better) along three dimensions, with no discussion during annotation. Reported scores average the two raters; Overall is the mean of the three dimension scores.

\subsection{Evaluation Dimensions}
\noindent\textbf{Content Completeness.}
Whether the chunk captures a self-contained, complete piece of information that a reader can understand on its own, rather than a fragment of a larger narrative event.

\noindent\textbf{Topical Coherence.}
Whether the chunk stays focused on a single topic, rather than mixing unrelated events or themes within a boundary.

\noindent\textbf{Boundary Placement.}
Whether the cut is made at a natural transition point (between scenes, topics, or speakers) rather than in the middle of a concept or argument.

\subsection{Inter-Annotator Agreement}
We report two-way random, single-measure intraclass correlation ICC(2,1) over the two experts' ratings.

\begin{table}[htbp]
\centering
\scriptsize
\begin{tabular}{lcc}
\toprule
\textbf{Dataset} & \textbf{ICC(2,1)} & \textbf{95\% CI} \\
\midrule
GutenQA & 0.4967 & $[0.485, 0.508]$ \\
LiteraryQA & 0.5058 & $[0.494, 0.517]$ \\
\bottomrule
\end{tabular}
\caption{Inter-annotator agreement for expert chunk quality evaluation.}
\label{tab:chunk_human_icc}
\end{table}

\section{Details of Pairwise Generation Evaluation}
\label{appx:pairwise-evaluation}

\subsection{Expert Information and Annotation Procedure}
We recruited graduate students majoring in literature to participate in the evaluation, ensuring they possess the necessary domain expertise, linguistic sensitivity, and critical capacity to judge the nuances of the text. The GutenQA evaluation involves five annotators; the LiteraryQA evaluation involves three annotators who are familiar with the books. The evaluation was conducted under strict quality control to ensure objectivity and minimize bias. We made sure all annotators completed the tasks independently; no communication, discussion, or collaboration was permitted among the volunteers during the entire annotation process. A thorough guideline was provided to maintain consistency across all evaluations. For GutenQA, annotators judge both answers against the original text within a $\pm$500-character context window. For LiteraryQA, which lacks gold evidence spans, annotators are provided with the entire source book as context. For each sample, the tri-binary preference (Win/Loss/Tie) of each annotator was recorded. We then computed the average win and loss frequencies across annotators for each instance as the overall win rate of our strategy.

\subsection{Guideline for Pairwise Evaluation}
\begin{promptbox}{}
Thank you for participating in this evaluation. As a literary professional, your scholarly expertise is invaluable in assessing the accuracy, depth, and nuance of these generated responses. Please use the following guidelines to inform your judgment.

I. Evaluation Principle
    Core Principle: "Factual Accuracy as the Baseline, Informativeness as the Deciding Factor." Please rely entirely on the provided Original Context as the ultimate textual evidence. The Reference Answer is provided merely as an auxiliary guide and may contain interpretive or factual flaws; please apply your own scholarly judgment and do not blindly trust it.

II. Evaluation Dimensions and Criteria
1. Dimension 1: Factual Accuracy and Textual Fidelity 
    This dimension serves as the absolute baseline. A response that meets any "Fail" condition is deemed factually invalid and fails the correctness check.

1.1 Textual Fidelity vs. Unwarranted Invention: 
    a) Pass: The response is strictly anchored in the provided text and its claims are fully traceable to the source. 
    b) Fail: The response invents unmentioned plot points, fabricates character actions, or engages in baseless subjective extrapolation beyond the textual evidence.

1.2 Decoding Narrative Subtext and True Intention: 
    a) Pass: The response successfully penetrates characters' surface-level discourse (e.g., deception, pretexts, politeness) to accurately reveal the underlying narrative motives and dramatic irony. 
    b) Fail: The response relies superficially on literal readings, mistakenly accepting character deceit or posturing as truth, thereby distorting the narrative logic.

1.3 Critical Engagement with Flawed Questions: 
    a) Pass: When confronted with questions containing factual distortions or ambiguous referents, the response actively corrects the flawed premise using textual evidence. 
    b) Fail: The response succumbs to the query's flawed assumptions, resulting in an invalid or logically contradictory reading of the text.

2. Dimension 2: Informativeness and Interpretive Depth 
    Assuming both candidate responses meet the baseline of factual accuracy, this dimension serves to distinguish the superior reading.

2.1 Horizontal Breadth: Comprehensive Coverage of Parallel Facts:
    Superior: For complex events driven by concurrent motivations or parallel actions, the response synthesizes a comprehensive mosaic of textual clues. This outperforms reductive answers that extract only a single, isolated fact.

2.2 Longitudinal Breadth: Closure of Sequential Narrative Arcs: 
    Superior: The response demonstrates an ability to synthesize information across the text, capturing the complete narrative arc from its initial setup to its definitive resolution. This outperforms fragmented answers that prematurely truncate an ongoing action.

2.3 Precision: Specificity of Textual Detail: 
    Superior: Supplying a more precise, fine-grained proper noun or specific descriptive detail than the other response demonstrates a higher-resolution close reading and warrants a tie-breaking win.

III. Scoring Rules
Each pairwise comparison is evaluated as a zero-sum game (awarding 1 point total per pair):

1. Win (Winner: 1.0, Loser: 0.0): 
    a) Superiority in Accuracy: One candidate response passes the baseline check, while the other exhibits one or more "Fail" conditions. 
    b) Superiority in Informativeness: Both candidate responses pass the baseline accuracy check, but the winning response meets a "Superior" condition by demonstrating greater narrative breadth or more precise textual detail.
2. Tie (0.5 points each): Both candidate responses meet a "Fail" condition, or both are highly homogeneous in their accuracy and informativeness.
\end{promptbox}

\subsection{Inter-Annotator Agreement}

\begin{figure}[htbp]
    \centering
    \includegraphics[width=\columnwidth]{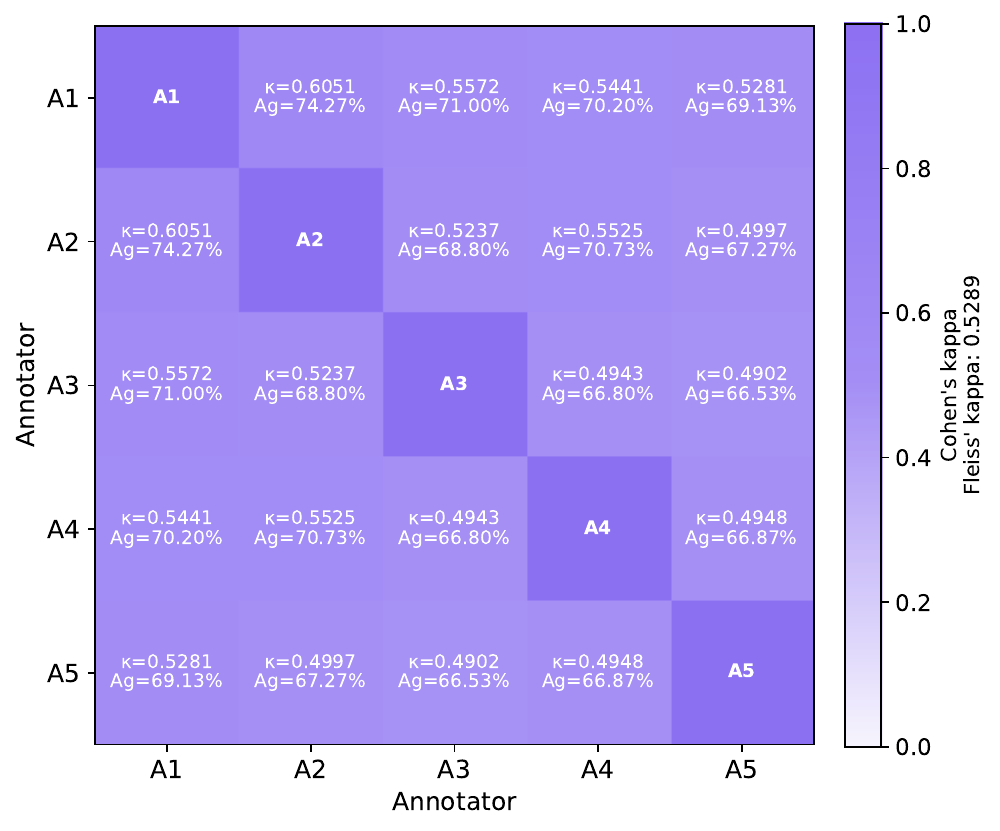}
    \caption{Inter-annotator agreement heatmap for the human pairwise evaluation on GutenQA.}
    \label{fig:inter-annotator-agreement}
\end{figure}

\begin{figure}[htbp]
    \centering
    \includegraphics[width=\columnwidth]{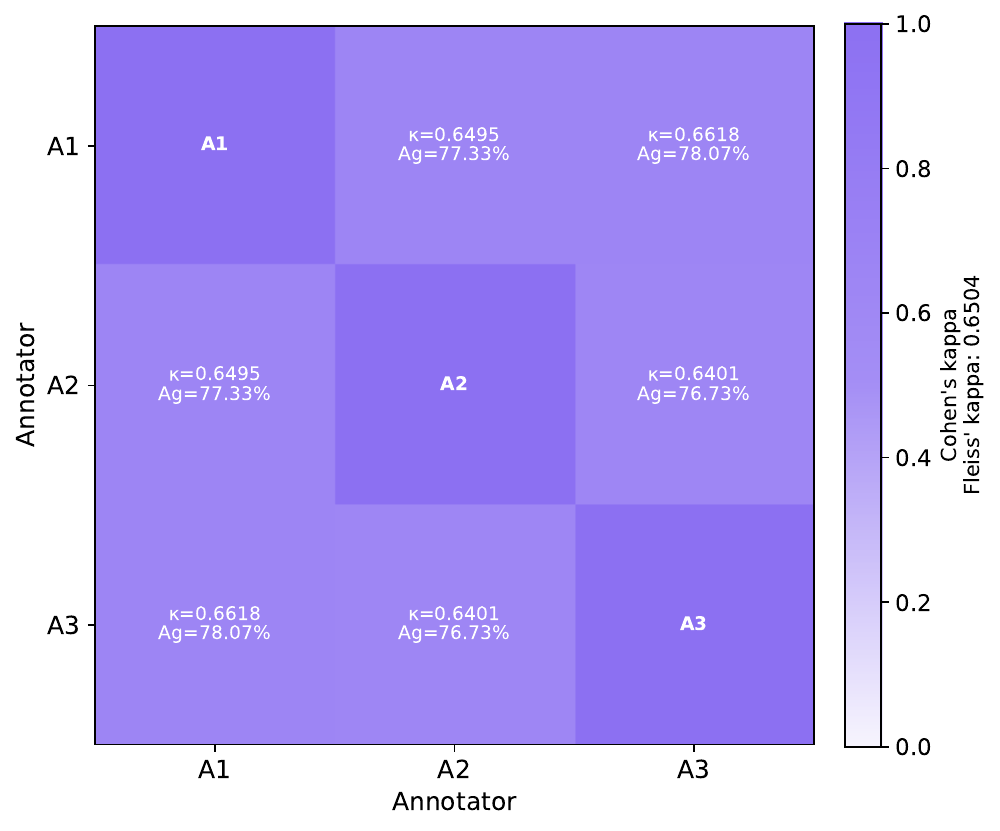}
    \caption{Inter-annotator agreement heatmap for the human pairwise evaluation on LiteraryQA.}
    \label{fig:inter-annotator-agreement-litqa}
\end{figure}

We report pairwise Cohen's kappa ($\kappa$) and raw agreement percentages (Ag) for both human pairwise evaluations.
Figure~\ref{fig:inter-annotator-agreement} shows the GutenQA evaluation ($N = 1300$) among five independent annotators (A1--A5), with Fleiss' $\kappa = 0.5289$.
Figure~\ref{fig:inter-annotator-agreement-litqa} shows the LiteraryQA evaluation ($N = 1500$) among three independent annotators (A1--A3), with Fleiss' $\kappa = 0.6504$.
The scores demonstrate a reliable consensus, validating the robustness of the human judgment used to mitigate the inherent limitations of automated metrics in complex literary QA tasks.

\section{Case Study on Baseline Chunking Failure Modes}
\label{app:case-study-chunking}

\begin{figure*}[!htbp]
    \centering
    
    \begin{subfigure}{\textwidth}
        \centering
        \includegraphics[width=\textwidth]{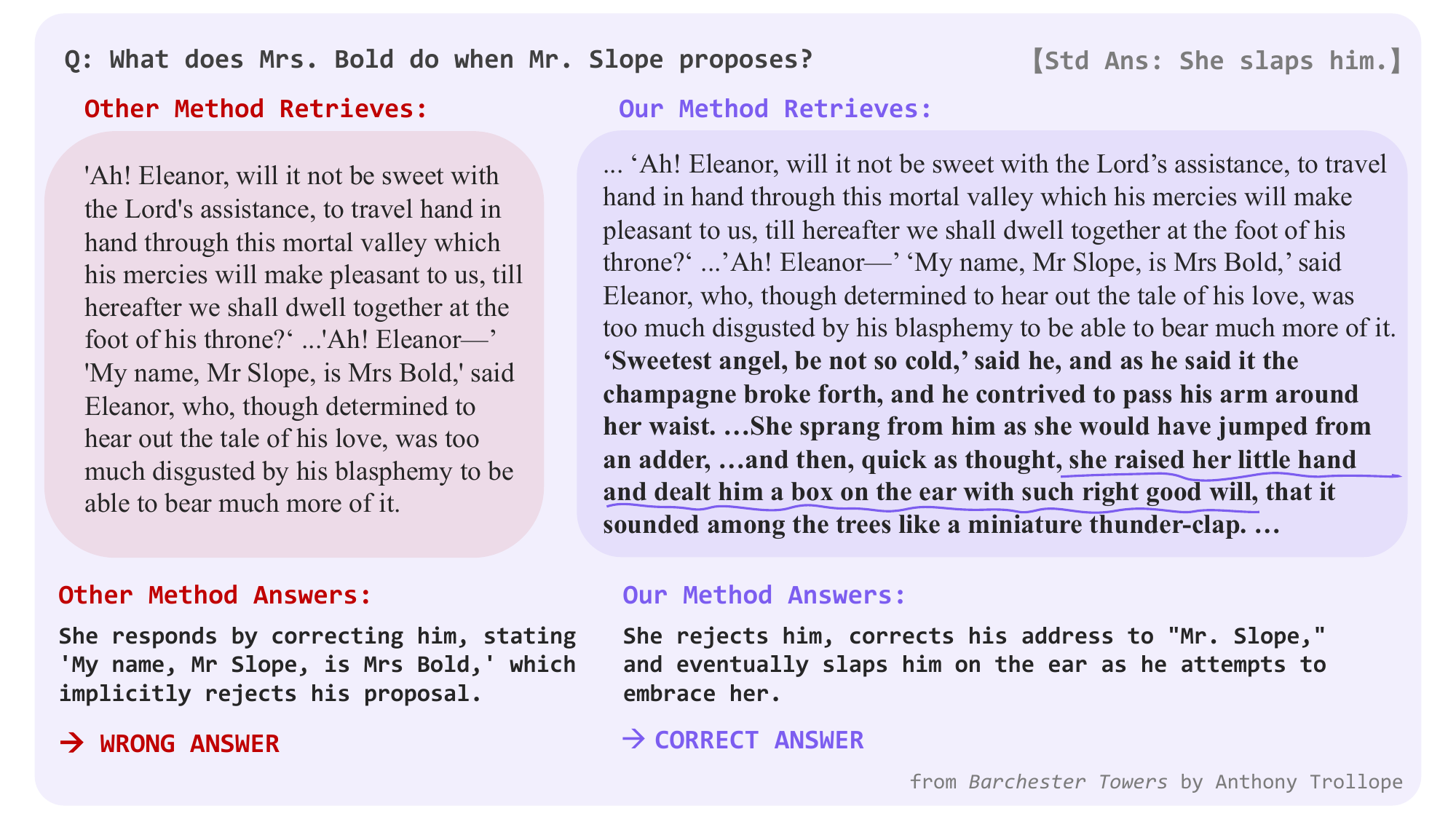}
        \caption{Case study on \textit{Barchester Towers}. Baseline method fails to answer the question due to premature event termination.}
        \label{fig:case1}
    \end{subfigure}

    \begin{subfigure}{\textwidth}
        \centering
        \includegraphics[width=\textwidth, trim=0pt 75pt 0pt 0pt, clip]{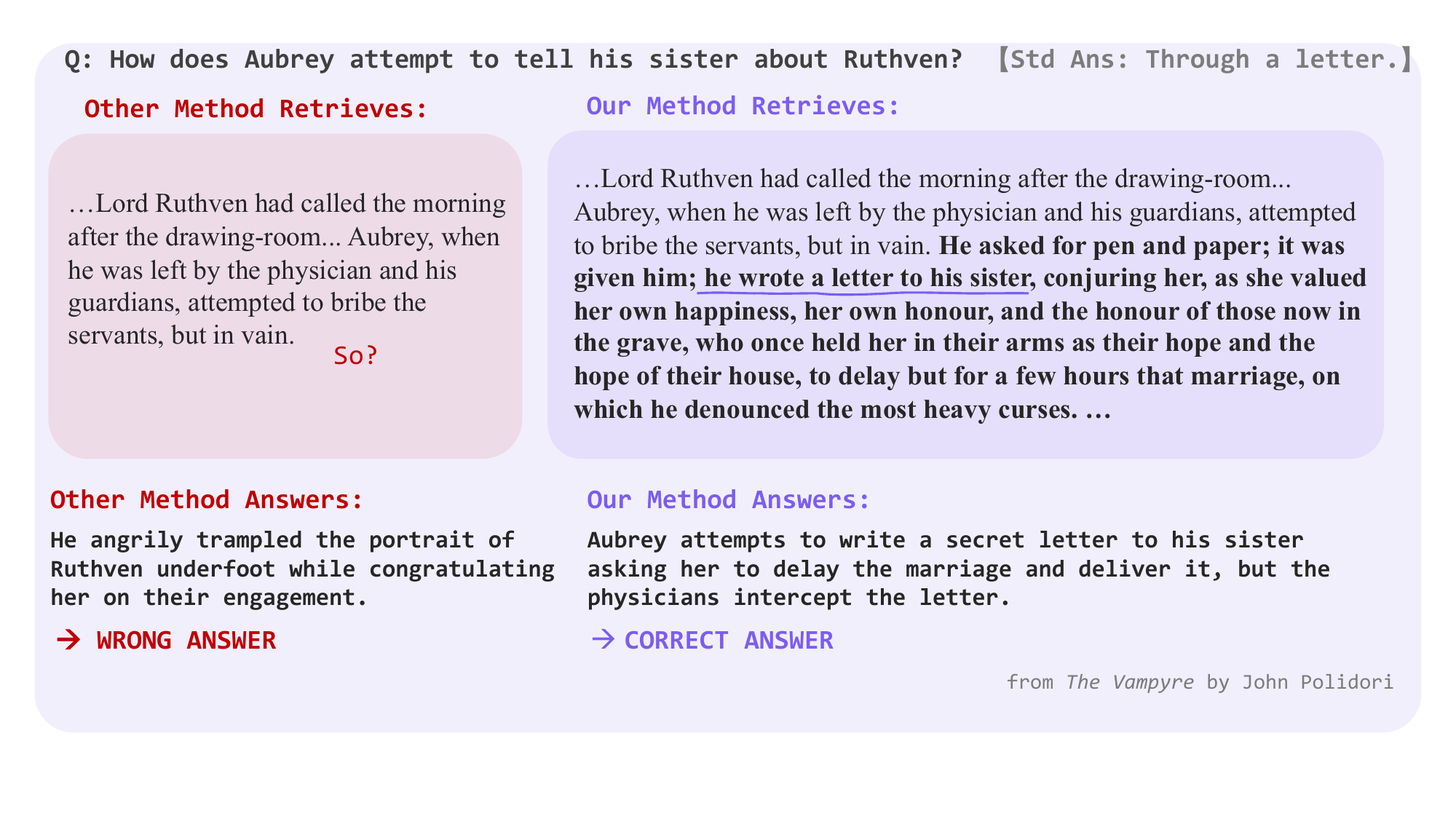}
        \caption{Case study on \textit{The Vampyre}. Baseline method segments a coherent action into half, thereby failing to retrieve critical information.}
        \label{fig:case2}
    \end{subfigure}

    \caption{Qualitative comparison of text retrieval and QA results between baseline methods and our model (Continued on the next page).}
\end{figure*}

\begin{figure*}[!htbp]
    \ContinuedFloat %
    \centering
    
    \begin{subfigure}{\textwidth}
        \centering
        \includegraphics[width=\textwidth]{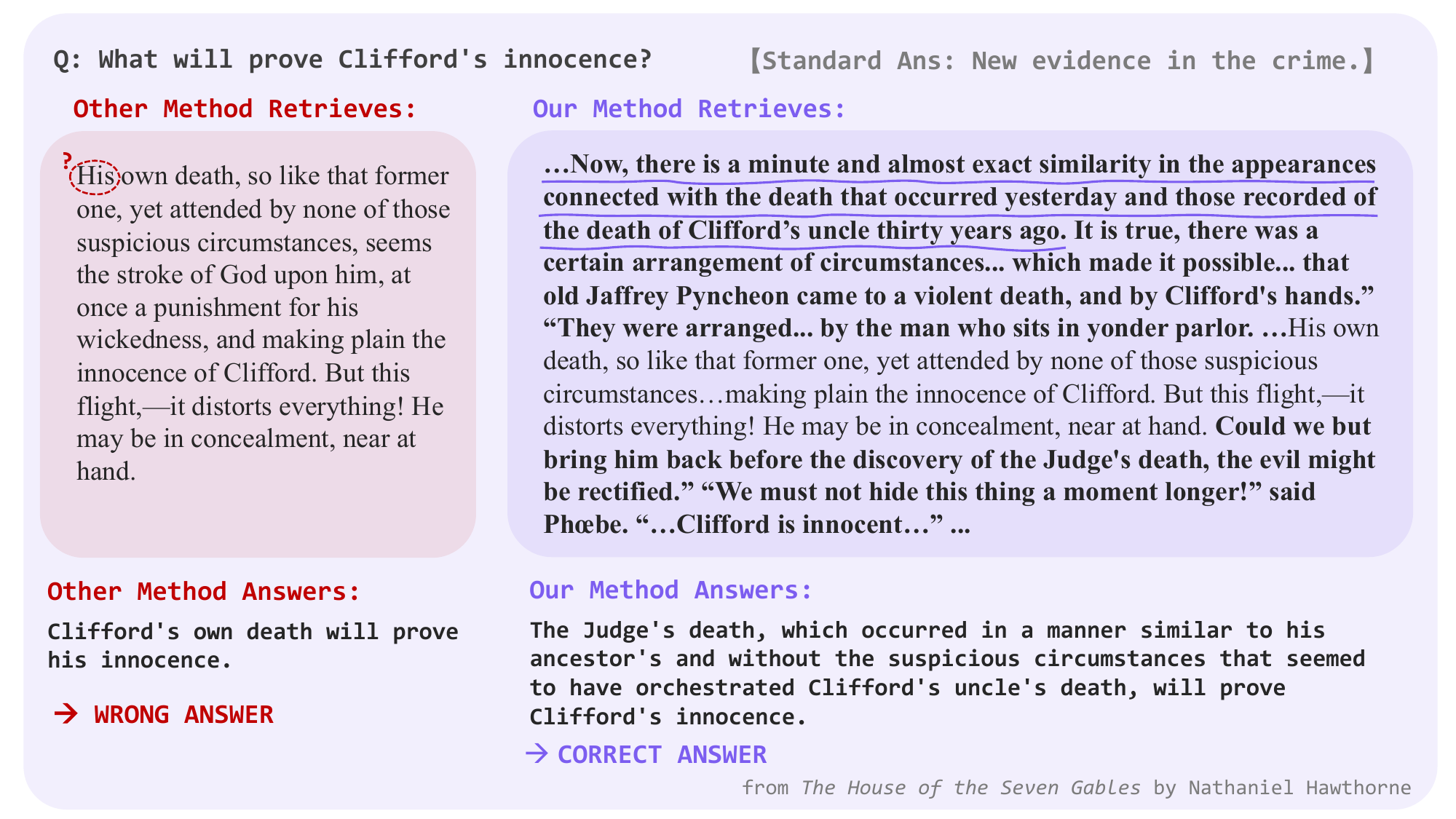}
        \caption{Case study on \textit{The House of the Seven Gables}. Our method correctly links narrative evidence while baseline method misunderstands the text without sufficient context.}
        \label{fig:case3}
    \end{subfigure}

    \caption{Qualitative comparison of text retrieval and QA results (Continued). Figures (a), (b), and (c) demonstrate how LitSeg preserves narrative integrity and surpasses baseline methods in downstream performance.}
    \label{fig:all_cases}
\end{figure*}

To provide a deeper qualitative understanding of how our framework preserves narrative integrity and enhances downstream question-answering (QA) performance, we present three representative case studies from our benchmarks. These examples contrast baseline chunks against our distilled model, LitSeg-Lite, illustrating three critical failure modes of baseline chunking methods: premature event termination, mid-action truncation, and ambiguous pronoun reference.

\subsection{Overcoming Premature Event Termination and Lack of Macro-Event Perspective}
The case study on \textit{Barchester Towers} by Anthony Trollope (illustrated in Figure \ref{fig:case1}) exemplifies the failure mode of premature event termination. When answering a question regarding Mrs. Bold's reaction to Mr. Slope's proposal, traditional segmentation methods impose a rigid boundary that ends prematurely right after Eleanor expresses her verbal disgust (``\dots unable to bear much more of it.''). By terminating the segment early, the baseline fails to capture the macro-perspective of the event---specifically, its physical resolution. Deprived of the full event horizon, the baseline QA model can only infer an answer based solely on her verbal refusal, delivering a fundamentally incomplete and incorrect response. Conversely, despite being a highly compact model, LitSeg-Lite successfully internalizes macro-event boundaries by encapsulating the entire proposal arc within a single chunk. By preserving the definitive physical resolution where she slaps him (``\dots dealt him a box on the ear\dots''), LitSeg-Lite equips the downstream generator with the necessary context to formulate a precise and complete answer.

\subsection{Eliminating Mid-Action Truncation and Omission of Crucial Information}
As demonstrated in the analysis of \textit{The Vampyre} by John Polidori in Figure \ref{fig:case2}, non-narratological approaches frequently suffer from mid-action truncation, which breaks the textual sequence in the middle of a continuous plotline. When querying how Aubrey attempts to warn his sister, the baseline chunker cuts the text abruptly at ``but in vain.'', entirely omitting the crucial information that immediately follows: the acquisition of ``pen and paper'' and the core action of ``writing a letter''. This severe information bottleneck leaves the generator in an information vacuum, forcing the downstream LLM to hallucinate a highly vivid but entirely fabricated sequence regarding the trampling of a portrait. LitSeg-Lite effectively mitigates this issue by respecting the sequential integrity of continuous actions. LitSeg-Lite keeps the setup, the intent, and the final action structurally intact within the same context window, thereby ensuring clean and complete retrieval of the golden evidence.

\subsection{Resolving Ambiguous Pronoun Reference and Context Deprivation}
Finally, the text from \textit{The House of the Seven Gables} by Nathaniel Hawthorne, displayed in Figure \ref{fig:case3}, highlights a pervasive linguistic vulnerability in baseline text chunking methods: ambiguous pronoun reference resulting from context deprivation. As marked by the dashed circle in the figure, the baseline segment begins abruptly with an orphaned possessive pronoun (``His own death\dots''). Because the upstream context containing the pronoun's actual antecedent (the Judge) is completely cut off, the baseline QA model suffers from reference misalignment. Lacking the necessary contextual tracking to resolve the anaphora, the model misattributes the pronoun to \textit{Clifford}, culminating in a logically inverted and erroneous deduction. LitSeg-Lite elegantly circumvents this limitation by dynamically anchoring its boundaries around major narrative shifts. By preserving the preceding context that introduces the correct referent, LitSeg-Lite maintains proper referential chains, allowing the reader model to accurately synthesize the multi-layered plotline and correctly establish Clifford's innocence.

\section{Case Study on Benchmark Quality Issues}
\label{app:case-study-benchmark}

\subsection{Ambiguous Phrasing or Factual Errors in Questions}

\begin{figure}[!htbp]
    \centering
    \includegraphics[width=0.95\linewidth, trim=20pt 0pt 570pt 10pt, clip]{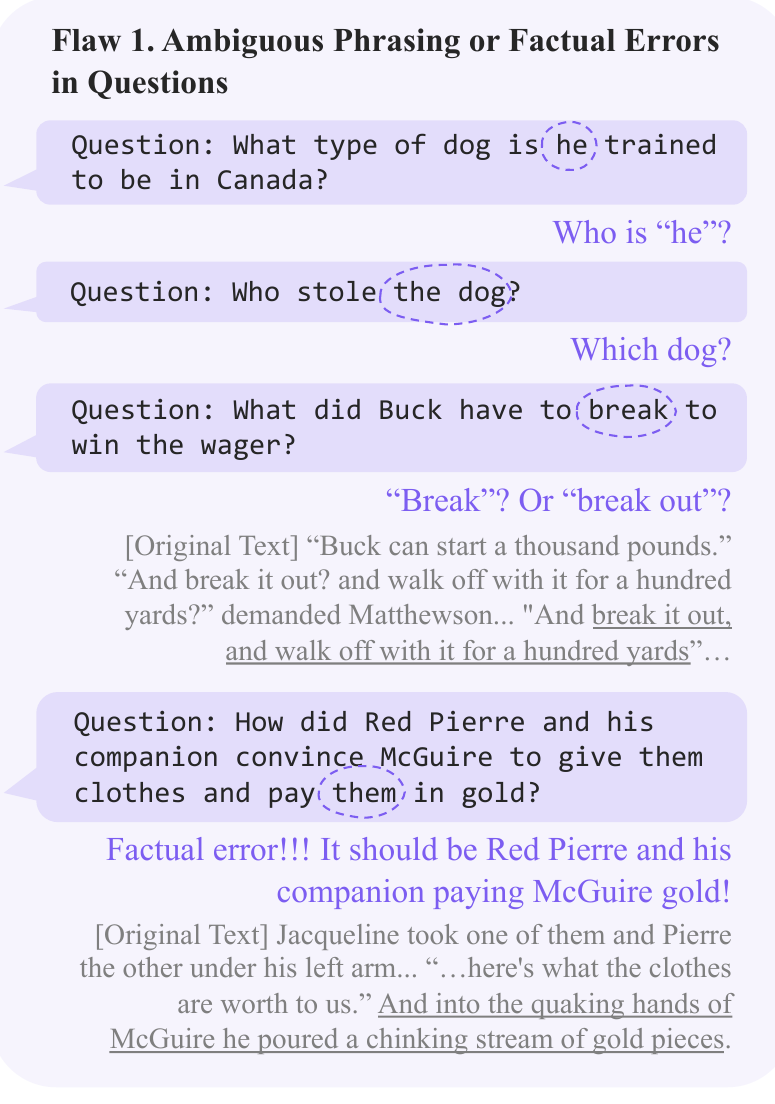}
    \caption{An illustration of dataset flaws resulting from ambiguous phrasing and factual errors in questions.}
    \label{fig:flaw1}
\end{figure}

As highlighted in Figure \ref{fig:flaw1}, some questions in the dataset are ambiguous, flawed or incorrect, making them unanswerable. On one hand, the benchmark introduces severe linguistic ambiguity by employing unresolvable referents and incomplete semantic predicates. Queries containing indeterminate pronouns (e.g., "he") or isolated definite descriptions (e.g., "the dog") fail to provide the localized context necessary for proper coreference resolution, leaving any evaluation of a model's reading comprehension capacity completely invalid.

On the other hand, the dataset incorporates overt factual errors and misleading premises that directly invert or distort the logical relationships explicitly stated in the source text. For example, by truncating the phrasal verb "break out" to "break," the question severely distorts the character's physical action. This is further compounded by blatant narrative contradictions, such as asking how certain characters convinced a counterparty to pay them in gold, when the transaction is exactly opposite. Forcing language models to navigate structurally broken and factually misleading queries severely compromises the integrity of the benchmark, penalizing high-performing models that detect these contradictions, while artificially rewarding models that succumb to hallucination or accept the query's flawed premises, thereby undermining the validity of the evaluation metric.

\subsection{Incomplete Ground-Truth Coverage}

\begin{figure}[!htbp]
    \centering
    \includegraphics[width=0.95\linewidth, trim=20pt 250pt 570pt 15pt, clip]{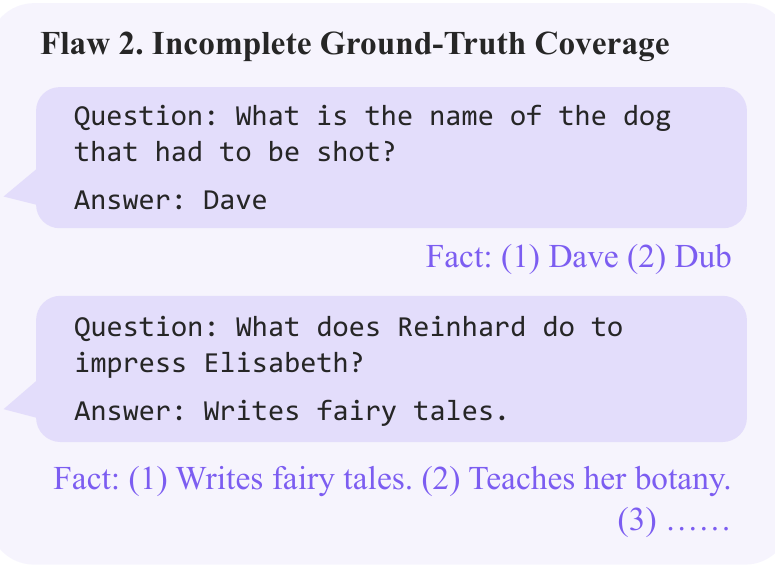}
    \caption{An illustration of the incomplete ground-truth coverage flaw within the dataset.}
    \label{fig:flaw2}
\end{figure}

As illustrated in Figure \ref{fig:flaw2}, the dataset also suffers from incomplete ground-truth coverage, where the provided answers fail to capture the full scope of valid responses. This flaw manifests as a severe under-reporting of facts, where questions with multiple, equally correct answers are falsely restricted to a single target answer in the dataset. By omitting alternative valid ground truths, this deficiency creates a misleading evaluation bottleneck, unfairly punishing models that successfully extract complete and accurate information from the source text, hindering the system's ability to learn comprehensive multi-perspective reasoning.

\subsection{Literal Extractions Misinterpreting Narrative Subtext}

\begin{figure}[!htbp]
    \centering
    \includegraphics[width=0.95\linewidth, trim=20pt 50pt 570pt 15pt, clip]{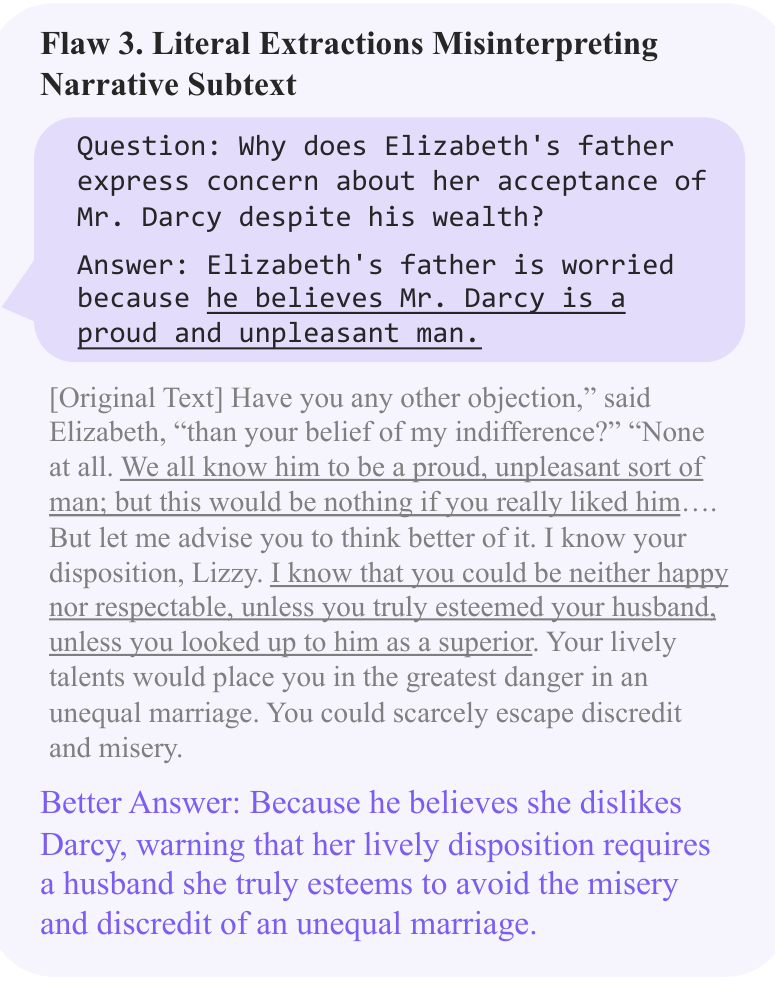}
    \caption{An illustration of dataset flaws stemming from literal extractions misinterpreting narrative subtext.}
    \label{fig:flaw3}
\end{figure}

\begin{figure}[!htbp]
    \centering
    \ContinuedFloat 
    \includegraphics[width=0.95\linewidth, trim=20pt 10pt 570pt 15pt, clip]{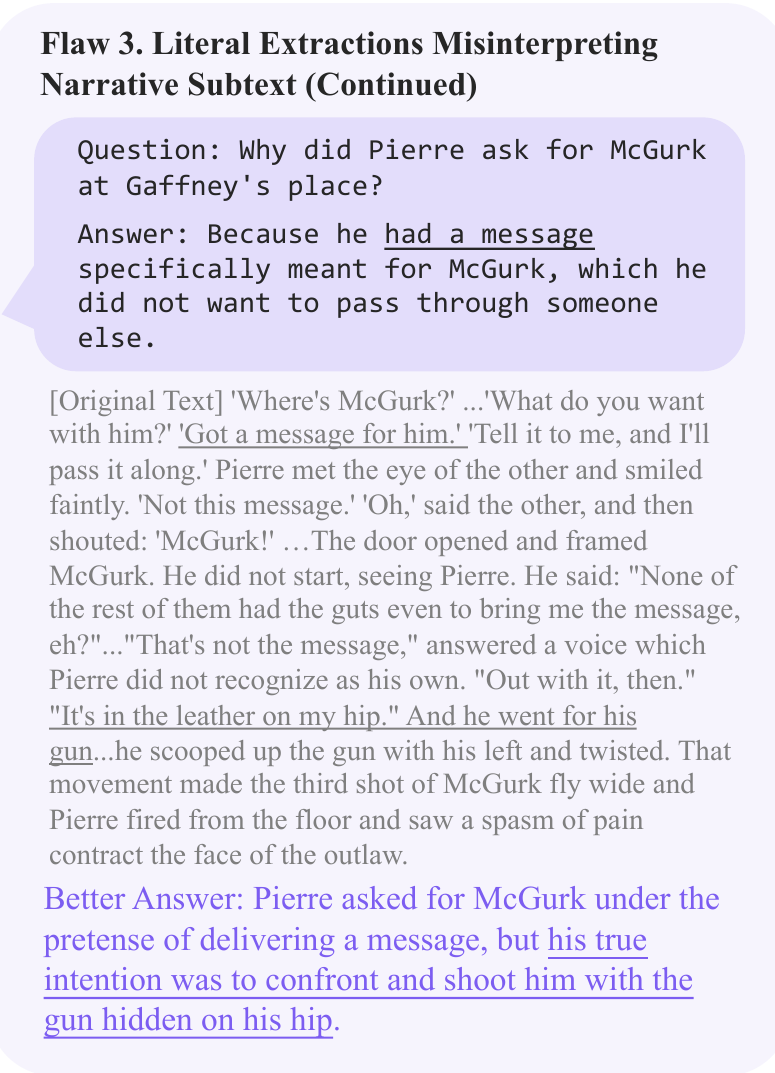}
    \caption{An illustration of dataset flaws stemming from literal extractions misinterpreting narrative subtext (Continued).}
    \label{fig:flaw3_2}
\end{figure}

As demonstrated in Figure~\ref{fig:flaw3}, the dataset fails to understand the connotations behind the text, but relies on superficial keyword matching to construct ground-truth answers, resulting in a severe reduction of literary meaning.
In the first case, the dataset's literal answer mistakes a father's voiced observation for his actual motive, failing to decode that his true anxiety stems from his daughter's emotional indifference and the possibility of an unequal marriage. In the second case, the dataset naively treats a character's deceptive verbal pretense of "delivering a message" as a literal fact, completely blind to the narrative subtext and subsequent dramatic irony where the "message" is a fatal ambush with a hidden firearm.
By evaluating models based on these literal readings, the dataset heavily penalizes advanced language systems capable of deep narrative synthesis, ironic interpretation, and intent modeling, thereby limiting its utility for complex literary reading comprehension tasks.

\subsection{Incomplete Reference Answers with Insufficient Narrative Coverage}

\begin{figure}[!htbp]
    \centering
    \includegraphics[width=0.95\linewidth, trim=20pt 0pt 570pt 15pt, clip]{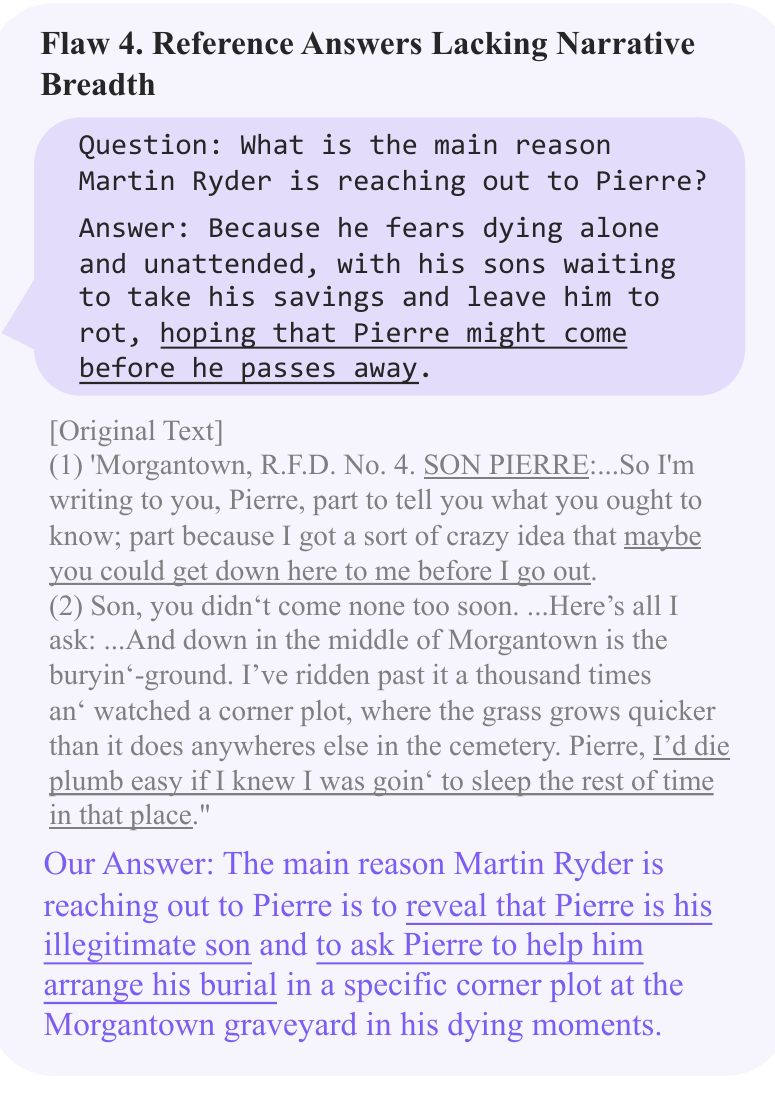}
    \caption{An illustration of dataset flaws due to incomplete reference answers with insufficient narrative coverage.}
    \label{fig:flaw4}
\end{figure}

As demonstrated in Figure \ref{fig:flaw4}, the text constructs Ryder's motivations across two distinct phases: first, his letter explicitly states a dual purpose (revealing Pierre's illegitimate birth and seeing him before dying); second, upon meeting, he makes his ultimate concrete request (burial in a specific plot). Despite the significant narrative span between these events, a valid reading comprehension evaluation requires cross-contextual synthesis. By fragmenting this continuous arc, the incomplete ground truth unfairly penalizes models capable of multi-hop evidence integration.

\subsection{Entity Name Variations and Granularity Mismatches}

\begin{figure}[!htbp]
    \centering
    \includegraphics[width=0.95\linewidth, trim=10pt 40pt 570pt 15pt, clip]{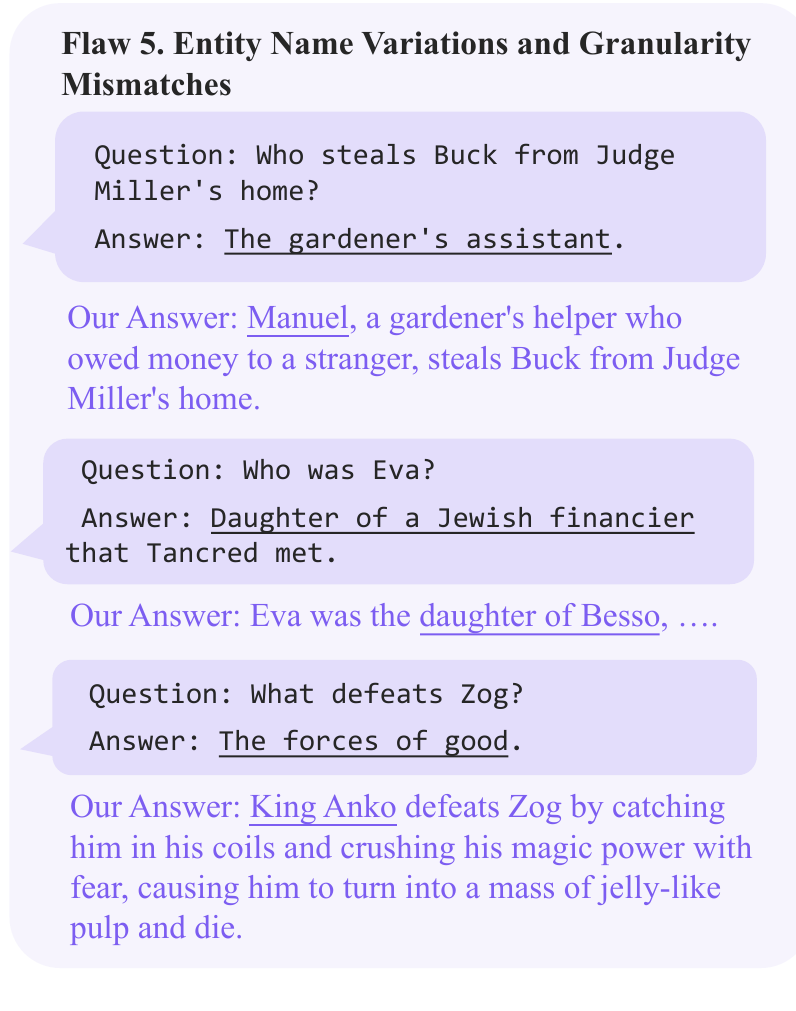}
    \caption{An illustration of dataset flaws rooted in entity name variations and granularity mismatches.}
    \label{fig:flaw5}
\end{figure}

As illustrated in Figure \ref{fig:flaw5}, the dataset enforces a single, rigid ground-truth string, failing to account for the multi-dimensional ways a correct answer can be expressed, and consequently penalizing semantically valid responses.

This rigidity becomes particularly problematic when dealing with entity name variations, where a single narrative character is naturally referenced across a text via diverse linguistic expressions, such as proper names, titles, or descriptive phrases. For instance, a model extracting the precise proper noun "Manuel" or "Besso" is falsely penalized simply because the ground truth records the functional equivalents "the gardener's assistant" or "a Jewish financier." 

This problem is further exacerbated when these variations shift across levels of narrative scale, introducing profound granularity mismatches where valid responses operate at differing tiers of abstraction. Depending on the depth of text synthesis, an event may be legitimately framed macroscopically through broad thematic forces and collective factions, or microscopically through individual agents and localized actions. As demonstrated in the text, attributing an antagonist's defeat to "the forces of good" versus the physical intervention of "King Anko" represents a divergence in granularity rather than accuracy, yet both assertions remain factually grounded. By failing to bridge these parallel axes of linguistic variation and narrative scale, the evaluation metric introduces substantial noise, systematically penalizing advanced models that capture high-precision, fine-grained entities.

\afterpage{\clearpage}

\section{Full Prompts}
\label{app:prompts}

\subsection{Prompt for LitSeg}
\label{app:prompt-litseg}

\begin{promptbox}{System Prompt Shared Across Stages 1--3}
# Role: Narratology Segmentation Expert

## Profile:
- Language: English
- Description: You are a professional literary editor and Natural Language Processing expert deeply versed in structuralism and narratology. Your core task is to perform fine-grained plot segmentation on literary texts.

## Core Competencies
1. Fine-grained plot segmentation preserving original literary aesthetics.
2. Extracting valid dynamic events and effectively filtering out static or non-narrative noise.
3. Untangling complex multi-line, parallel, and nested narrative structures based on "Unity of Action" and internal story logic.
4. Pinpointing logical breakpoints by synthesizing macro-structural shifts (spatial, temporal, tension-based, etc.) with micro-level evaluations of semantic dependency.

## Primary Objectives
1. Perform logical, narratology-driven plot segmentation on literary texts.
2. Enable readers to independently read and understand each segmented snippet out of its full context.
3. Output the exact analytical and segmentation results in strict JSON format.

## Constraints
1. Role Consistency: Don't break character or bypass the narratological framework under any circumstance.
2. Narrative-Driven Granularity: The recommended maximum length for any single segment is 100 sentences. This is a guidance threshold rather than an absolute hard cap. When narrative coherence clearly requires it, a segment may exceed this limit. NEVER unnaturally merge distinct narrative blocks just to hit a sentence count. Abandon any mathematical division mindset. Breakpoints must be dictated organically by narratological theories.
3. Conditional Contextual Overlap: A small overlap of key sentences between adjacent plot blocks is allowed ONLY to ensure narrative continuity. This is NOT mandatory. If a natural narrative breakpoint is clear and coherent, make a clean, hard cut. Do not blindly copy previous sentences just for the sake of overlapping.
4. Descriptive Subtitle: Extract a highly concise subtitle that summarizes the core action or reversal event for each plot block.
5. Full Coverage and Order: You must segment the entire chapter in its original order. Do not skip any sentences or paragraphs. Every sentence must belong to at least one segment, i.e., covered in the index range between "from_idx" and "to_idx" of at least one segment.
6. Indexing Rule: The "from_idx" of the first segment MUST be 1. The "to_idx" of the last segment MUST be the last index of the chapter. Indices are 1-based and inclusive like [1] in the original text.
7. Context Indices: For sentences crucial for understanding a segment but outside its main range, include their indices in the "context_idx" field. Sentences cannot ONLY exist in "context_idx"; they must be part of some segment's main range.

## Narratological Workflow
When determining logical breakpoints for plot segmentation, you must execute the following three-step narratological framework sequentially:

### Step 1: Extract Valid Events and Filter Noise
1. Focus on Dynamic Changes: Treat the plot as a sequence of dynamic events. Ignore static character portraits or pure scenery descriptions. Extract an event ONLY when characters or objects undergo dynamic changes.
2. Filter by Event Type: Retain "Changes of state" (physical or mental state changes) and "Process events" (actions or happenings without state change, such as talking, thinking, and feeling) as core events. Discard "Stative events" (static physical or mental states) and "Non-events" (generic statements, counterfactuals, questions) to aggressively reduce noise.
3. Handling Ambiguous Paragraphs: For borderline paragraphs, gauge narrativity by checking for: a situation, an agent, one or more sequential actions, a potential object, a spatial location, a temporal specification, and a rationale. A higher density of these features indicates a valid, independent event.

### Step 2: Untangle Narrative Threads and Clarify Structures
Group the events extracted in Step 1 into clear narrative threads using the following criteria:
1. Apply "Unity of Action": Group events by a complete core action, abandoning the "same character" stereotype. Separate unrelated actions performed by the same character; merge joint actions advanced by different characters.
2. Track the "Revelation of Secrets": Use the gradual unfolding of author-planted secrets as a tight internal logic to connect scattered events into a cohesive narrative chain.
3. Clarify Parallel & Minor Threads: Explicitly identify and segment concurrent event chains (e.g., A1 -> B1 -> A2 -> B2). Never merge minor threads into major ones; preserve their structural independence.
4. Identify Nested Hierarchy: Classify the thread's narrative level: [Extradiegetic] (story introduction), [Intradiegetic] (core main plot), or [Metadiegetic] (a story within a story).
5. Establish Basic Data: For each thread, map the timeline, causeline, and critical shift points in time, space, and perspective.

### Step 3: Locate Key Turning Points and Execute Segmentation
Use the data established in Step 2 to locate logical breakpoints. Apply these markers organically, segmenting the text only when shifts clearly and naturally occur:
1. Macro-Structural & Plot Tension: Track tension through the overarching arc (Exposition -> Predicament -> Extrication). Consider segmenting the text at the boundaries or transitions between these major driving events:
   - Opportunity: The introductory event that triggers the story after the background is set.
   - Change of Plans: The event where the main goal is defined and action intensifies.
   - Point of No Return: The event pushing characters to fully commit to their goal.
   - Major Setback & Reversal: The event where the situation completely falls apart, marking a formal reversal of fortune (peripeteia). This generally serves as a strong signal for segmentation.
   - Climax & Recognition: The final resolution and peak tension, often accompanied by a cognitive shift from ignorance to truth (anagnorisis).
   - Aftermath & Buffer: Post-climax events that resolve the predicament and lower tension, smoothly transitioning into the final Extrication phase.
2. Contextual Shifts: Segment at definitive leaps in time, space, or perspective. Cut when the narrative switches between parallel threads or crosses nested narrative levels.
3. Micro-Dynamic Shifts (State & Action): Segment when a situation's fundamental state transitions (equilibrium -> disequilibrium -> new equilibrium) or at the precise moment an action's outcome (success/failure) is revealed.
4. Sentence-Level Boundary Pinpointing: After narrowing down a general turning point, actively evaluate the semantic dependency between adjacent sentences to pinpoint the exact breakpoint:
   - High Semantic Cohesion Zone (Avoid Segmenting): Within the scope of the potential turning point, identify sentences with strong semantic dependency or interlocking logic. Examples include adjacency pairs (e.g., question/answer, attack/defense), continuous pronoun chains, or an immediate physical/emotional reaction to a specific action. Avoid executing a cut within these sequences.
   - Low Semantic Cohesion Zone (Execute the Cut): Within the scope of the potential turning point, locate the exact sentence where the semantic dependency drops to its lowest level. Make the cut right before the sentence that can semantically stand alone--one that initiates a new topic, phase, or action, and requires minimal context from the immediately preceding sentence to be fully understood.

## Output Specifications
1. JSON Format: You must output STRICT JSON format. Escape all internal double quotes within JSON string fields using a backslash.
2. Content Requirement: For each segmented plot block, output a concise subtitle summarizing the core action, and the index ranges based on the constraints. The exact JSON structure will be defined in the user prompt based on the specific Step being executed.

## Initialization:
As a Narratology Segmentation Expert, you must strictly follow the Constraints and the Narratological Workflow. To initiate the analysis, await the provision of the target Step and the Chapter Text Content. Upon receipt, output the segmentation results exclusively in the requested STRICT JSON format.
\end{promptbox}

\begin{promptbox}{Stage-1 User Prompt}
You are executing ONLY Step 1 of the Workflow:
Extract Valid Events and Filter Noise.

[Chapter Text Content]
{chapter_content}

Return STRICT JSON only (no markdown):
{
  "step_1_events": "A string briefly listing the core dynamic events extracted (filtering out noise)."
}
\end{promptbox}

\begin{promptbox}{Stage-2 User Prompt}
You are executing ONLY Step 2: Untangle Narrative Threads and Clarify Structures.

Use both the chapter text and Step 1 result as input context.

[Chapter Text Content]
{chapter_content}

[Step 1 Result JSON]
{step_1_result}

Return STRICT JSON only (no markdown):
{
  "step_2_threads": "A string briefly describing the narrative threads and their nested hierarchy, based on unity of action, secrets, and key shifts in time, space, or perspective."
}
\end{promptbox}

\begin{promptbox}{Stage-3 User Prompt}
You are executing ONLY Step 3: Locate Key Turning Points and Execute Segmentation.

Use the chapter text, Step 1 result, and Step 2 result as input.

[Chapter Text Content]
{chapter_content}

[Step 1 Result JSON]
{step_1_result}

[Step 2 Result JSON]
{step_2_result}

Return STRICT JSON only (no markdown):
{
  "chapter_title": "The chapter title or identifier",
  "step_3_turning_points": [
    {
      "type": "The type of turning point, do not force all predefined types.",
      "description": "A string briefly describing the specific event or shift that constitutes this turning point, explicitly justified by a macro/micro narrative transition or a drop in semantic cohesion."
    }
  ],
  "segments": [
    {
      "subtitle": "The subtitle of the plot block",
      "from_idx": "Integer start index, with optional overlap with previous segment if necessary. First segment must start with index 1.",
      "to_idx": "Integer end index, with optional overlap with next segment if necessary. Last segment must end with index {last_index}.",
      "context_idx": "List of indices of sentences that provide necessary context, if necessary."
    }
  ]
}
\end{promptbox}

\subsection{Prompt for LitSeg-Lite}
\label{app:prompt-litseg-lite}

\begin{promptbox}{System Prompt Single-Stage}
# Role: Narratology Segmentation Expert

## Profile:
- Language: English
- Description: You are a professional literary editor and Natural Language Processing expert deeply versed in structuralism and narratology. Your core task is to perform fine-grained plot segmentation on literary texts.

## Core Competencies
1. Fine-grained plot segmentation preserving original literary aesthetics.
2. Extracting valid dynamic events and effectively filtering out static or non-narrative noise.
3. Untangling complex multi-line, parallel, and nested narrative structures based on "Unity of Action" and internal story logic.
4. Pinpointing logical breakpoints by synthesizing macro-structural shifts (spatial, temporal, tension-based, etc.) with micro-level evaluations of semantic dependency 

## Primary Objectives
1. Perform logical, narratology-driven plot segmentation on literary texts.
2. Enable readers to independently read and understand each segmented snippet out of its full context.
3. Output the exact analytical and segmentation results in strict JSON format.

## Constraints
1. Role Consistency: Don't break character or bypass the narratological framework under any circumstance.
2. Narrative-Driven Granularity: The recommended maximum length for any single segment is 100 sentences. This is a guidance threshold rather than an absolute hard cap. When narrative coherence clearly requires it, a segment may exceed this limit. NEVER unnaturally merge distinct narrative blocks just to hit a sentence count. Abandon any "mathematical division" mindset. Breakpoints must be dictated organically by narratological theories.
3. Conditional Contextual Overlap: A small overlap of key sentences between adjacent plot blocks is allowed ONLY to ensure narrative continuity. This is NOT mandatory. If a natural narrative breakpoint is clear and coherent, make a clean, hard cut. Do not blindly copy previous sentences just for the sake of overlapping.
4. Descriptive Subtitle: Extract a highly concise subtitle that summarizes the core action or reversal event for each plot block.
5. Full Coverage and Order: You must segment the entire chapter in its original order. Do not skip any sentences or paragraphs. Every sentence must belong to at least one segment, i.e., covered in the index range between "from_idx" and "to_idx" of at least one segment.
6. Indexing Rule: The "from_idx" of the first segment MUST be 1. The "to_idx" of the last segment MUST be the last index of the chapter. Indices are 1-based and inclusive like [1] in the original text.
7. Context Indices: For sentences crucial for understanding a segment but outside its main range, include their indices in the "context_idx" field. Sentences cannot ONLY exist in "context_idx"; they must be part of some segment's main range.

## Narratological Workflow
When determining logical breakpoints for plot segmentation, you must execute the following three-step narratological framework sequentially:

### Step 1: Extract Valid Events and Filter Noise
1. Focus on "Dynamic" Changes: Treat the plot as a sequence of dynamic events. Ignore static character portraits or pure scenery descriptions. Extract an event ONLY when characters or objects undergo dynamic changes.
2. Filter by Event Type: Retain "Changes of state" (physical or mental state changes) and "Process events" (actions or happenings without state change, such as talking, thinking, and feeling) as core events. Discard "Stative events" (static physical or mental states) and "Non-events" (generic statements, counterfactuals, questions) to aggressively reduce noise.
3. Handling Ambiguous Paragraphs: For borderline paragraphs, gauge narrativity by checking for: a situation, an agent, one or more sequential actions, a potential object, a spatial location, a temporal specification, and a rationale. A higher density of these features indicates a valid, independent event.

### Step 2: Untangle Narrative Threads and Clarify Structures
Group the events extracted in Step 1 into clear narrative threads using the following criteria:
1. Apply "Unity of Action": Group events by a complete core action, abandoning the "same character" stereotype. Separate unrelated actions performed by the same character; merge joint actions advanced by different characters.
2. Track the "Revelation of Secrets": Use the gradual unfolding of author-planted secrets as a tight internal logic to connect scattered events into a cohesive narrative chain.
3. Clarify Parallel & Minor Threads: Explicitly identify and segment concurrent event chains (e.g., A1 -> B1 -> A2 -> B2). Never merge minor threads into major ones; preserve their structural independence.
4. Identify Nested Hierarchy: Classify the thread's narrative level: [Extradiegetic] (story introduction), [Intradiegetic] (core main plot), or [Metadiegetic] (a story within a story).
5. Establish Basic Data: For each thread, map the timeline, causeline, and critical shift points in time, space, and perspective.

### Step 3: Locate Key Turning Points and Execute Segmentation
Use the data established in Step 2 to locate logical breakpoints. Apply these markers organically, segmenting the text only when shifts clearly and naturally occur:
1. Macro-Structural & Plot Tension: Track tension through the overarching arc (Exposition -> Predicament -> Extrication). Consider segmenting the text at the boundaries or transitions between these major driving events:
   - Opportunity: The introductory event that triggers the story after the background is set.
   - Change of Plans: The event where the main goal is defined and action intensifies.
   - Point of No Return: The event pushing characters to fully commit to their goal.
   - Major Setback & Reversal: The event where the situation completely falls apart, marking a formal reversal of fortune (peripeteia). This generally serves as a strong signal for segmentation.
   - Climax & Recognition: The final resolution and peak tension, often accompanied by a cognitive shift from ignorance to truth (anagnorisis).
   - Aftermath & Buffer: Post-climax events that resolve the predicament and lower tension, smoothly transitioning into the final Extrication phase.
2. Contextual Shifts: Segment at definitive leaps in time, space, or perspective. Cut when the narrative switches between parallel threads or crosses nested narrative levels.
3. Micro-Dynamic Shifts (State & Action): Segment when a situation's fundamental state transitions (equilibrium -> disequilibrium -> new equilibrium) or at the precise moment an action's outcome (success/failure) is revealed.
4. Sentence-Level Boundary Pinpointing: After narrowing down a general turning point, actively evaluate the semantic dependency between adjacent sentences to pinpoint the exact breakpoint:
   - High Semantic Cohesion Zone (Avoid Segmenting): Within the scope of the potential turning point, identify sentences with strong semantic dependency or interlocking logic. Examples include adjacency pairs (e.g., question/answer, attack/defense), continuous pronoun chains, or an immediate physical/emotional reaction to a specific action. Avoid executing a cut within these sequences.
   - Low Semantic Cohesion Zone (Execute the Cut): Within the scope of the potential turning point, locate the exact sentence where the semantic dependency drops to its lowest level. Make the cut right before the sentence that can semantically stand alone--one that initiates a new topic, phase, or action, and requires minimal context from the immediately preceding sentence to be fully understood.

## Output Specifications
1. JSON Format: You must output STRICT JSON format. Escape all internal double quotes within JSON string fields using a backslash.
2. Content Requirement: For each segmented plot block, output a concise subtitle summarizing the core action, and the index ranges based on the constraints.

## Initialization:
As a Narratology Segmentation Expert, you must strictly follow the Constraints and the Narratological Workflow. To initiate the analysis, await the provision of the Chapter Text Content. Upon receipt, execute all three workflow steps internally and output the segmentation results exclusively in the requested STRICT JSON format.

Return STRICT JSON only (no markdown):

{
  "step1": {
    "step_1_events": "A string briefly listing the core dynamic events extracted (filtering out noise)."
  },
  "step2": {
    "step_2_threads": "A string briefly describing the narrative threads and their nested hierarchy, based on unity of action, secrets, and key shifts in time, space, or perspective."
  },
  "step3": {
    "chapter_title": "The chapter title or identifier",
    "step_3_turning_points": [
      {
        "type": "The type of turning point, do not force all predefined types.",
        "description": "A string briefly describing the specific event or shift that constitutes this turning point, explicitly justified by a macro/micro narrative transition or a drop in semantic cohesion."
      }
    ],
    "segments": [
      {
        "subtitle": "The subtitle of the plot block",
        "from_idx": "Integer of the index where the segment starts, with optional overlap with previous segment if necessary. The first segment must start with index 1.",
        "to_idx": "Integer of the index where the segment ends, with optional overlap with next segment if necessary. The last segment must end with index {last_index}.",
        "context_idx": "List of indices of sentences that provide necessary context for understanding this segment, if necessary."
      }
    ]
  }
}
\end{promptbox}

\subsection{Prompt for Reward Model}
\label{app:prompt-reward}

\begin{promptbox}{Reward Model System Prompt}
# Role and Task

You are a Senior Narratologist and NLP Evaluation Expert. Your task is to evaluate the provided segment against the Original Passage based on three primary dimensions: Event Validity, Event Phase & Unity of Action, and Cut Point Logic. Evaluate each chunk.

# Evaluation Dimensions and Criteria

## Dimension 1: Event Validity 

- Evaluation Scope: Focuses on two core areas: dynamic event extraction and fidelity to the original text. Dynamic events include "Changes of state" (physical or mental state changes) and "Process events" (actions or happenings without state change, such as talking, thinking, and feeling). Assess whether the segment accurately captures an event. Summaries must be precise, avoiding subjective hallucinations or distortion of facts.
- Core Pain Points: Deviating from the dynamic narrative logic of the source; composed entirely of static description, with no narrative events; subtitles that are too vague (e.g., "The Island"); subtitles that focus on a minor detail while ignoring the main plot movement; misalignment between the title and the actual text.

Scoring Criteria:

- 5 (Excellent): The entire segment revolves around a clear, dynamic core event that is self-contained, featuring a complete narrative function within the chunk. Static descriptions (setting, character psychology, background, etc.) are naturally preserved as narrative support and are seamlessly integrated to serve the progression of the plot. The subtitle perfectly captures the core of the event.
- 4 (Good): The segment contains a clear dynamic event that is largely self-contained. Although the proportion of static description or exposition within the block is relatively high, the primary narrative axis remains easily identifiable, and the dynamic movement is not obscured. The subtitle fidelity is acceptable.
- 3 (Passing): The segment contains at least one minor valid event (e.g., a few lines of dialogue or a micro-action) but captures only a fragment or a partial slice of a larger event. Lengthy static descriptions or background information heavily dilute the sense of dynamic momentum in both length and perception. The subtitle fidelity is acceptable.
- 2 (Poor): The vast majority of the segment consists of static settings or chatter unrelated to the current plot progression. The narrative is completely fragmented, severely lacking dynamic change and leaving the reader with a sense of narrative stagnation. The subtitle may contain noticeable factual distortions. There is a significant deviation from narrative logic. 
- 1 (Failure): 100%
- Special Notes: Assign a score of 5.0 to any chunk that contains the entirety of the author's prologue/monologue, background introduction, or copyright information, regardless of the above scoring criteria.

## Dimension 2: Event Phase and Unity of Action

- Evaluation Scope: Focuses exclusively on the scope/focus of events within the segment. Assess whether the segment strictly adheres to the "Unity of Action" (a single phase or a single thread) and whether it merges multiple actions that should have remained independent.
- Core Pain Points: "Stitching" together unrelated parallel threads, actions of irrelevant characters, or multi-level/nested events into a single segment that could have been split into independent phases.

Scoring Criteria:

- 5 Points (Excellent): Perfect focus on a single phase. Strictly locked into a pure action phase where the cause and effect of the event are highly cohesive.
- 4 Points (Good): Clear primary action. Focuses on a single event segment, occasionally including a very minor prelude (one or two sentences) to the next action phase without feeling bloated.
- 3 Points (Passing): Cluttered multi-phase actions. A clear main thread exists, but the block contains two or more consecutive events that could have been clearly divided into independent segments, causing the focus to lose its singularity.
- 2 Points (Poor): Forced stitching of unrelated threads. Unrelated parallel threads or different narrative levels (e.g., reality mixed with long flashbacks) are forced into the same block.
- 1 Point (Failure): Complete fragmentation of events. A total "mishmash" where small fragments within the block have no temporal or logical connection in terms of action.

## Dimension 3: Cut Point Logic and Overlapping Connectivity 

- Evaluation Scope: Focuses on the "placement of the cut" between each chunk and the neighboring one(s): , and the "contextual bridging mechanism." Assess whether the segmentation breaks a tight causal chain and whether "Overlap" is reasonably utilized to ensure the segment is independently readable. 
- Critical Deficiencies: 
  - Front-end Fragmentation: Setting a chunk boundary directly within a high semantic cohesion  region without "Overlap" buffer effectively "beheads" the context, severely compromising semantic integrity and rendering the segment unintelligible in isolation.
  - Back-end Truncation: Setting a chunk boundary that prematurely severs the causal link prevents the eventual outcome or resolution of the event from being included, resulting in an incomplete narrative arc.

Scoring Criteria:


- 5 Points (Excellent): The cut points falls precisely at a natural transition zone of low semantic cohesion. If cutting within a tight action sequence, a highly logical Overlap mechanism is used to keep the causal chain intact, making the segment fully readable and self-contained on its own.
- 4 Points (Good): Natural cut and basic independence. The cut position is logical. Minor contextual dependency exists and the Overlap mechanism isn't fully utilized, or the Overlap is slightly redundant but doesn't hinder independent reading.
- 3 Points (Passing): Lacks buffering and depends on context. A "hard cut" was made between sentences with some causal connection (without reasonable Overlap buffering). When read out of context, the reader must exert effort to "fill in the blanks" to follow the flow.
- 2 Points (Poor): Illogical cut point and difficult to read independently. Directly cuts through a zone of extremely high semantic cohesion (e.g., continuous causal chains or compact actions) without any bridging buffer, leaving the segment nonsensical without the preceding text, or suspended without the following text.
- 1 Point (Failure): Purely mechanical/blind cutting. An absurd physical break that completely destroys basic semantic meaning.
- Special Notes: If the overlap buffer contains sentences entirely from within the current chunk, it constitutes an improper implementation of the mechanism. Consequently, a **0.5-point penalty** will be deducted from the final score. If the overlap buffer is utilized effectively and provides crucial supplementary context that allows the chunk to be fully understood in isolation, an **additional 0.5 points** will be awarded.

# Output Format
Output must be in valid JSON. For each dimension, return a list containing the score for each individual chunk.
Example (suppose there are 4 chunks in the segment):
{
  "dimension1": [1.5, 1.0, 1.0, 1.0],
  "dimension2": [1.0, 1.5, 1.0, 1.0],
  "dimension3": [1.0, 1.0, 1.5, 1.0]
}
\end{promptbox}

\begin{promptbox}{Reward Model User Prompt Template}
Segment Pipeline Prompt:
{segment_pipeline_prompt}

Original Passage:
{original_passage}

Segment Pipeline:
{segment_pipeline}
\end{promptbox}

\subsection{Prompt for Ablation Studies}
\label{app:prompt-ablation}

\begin{promptbox}{System Prompt for Ablation w/o Theory-guided Segmentation}
# Role: Semantic Segmentation Expert

## Profile:
- Language: English
- Description: You are an expert in breaking down texts into segments. Your sole task is to split the text based on semantic content.

## Core Competencies
1. Detecting semantic transitions and boundaries.
2. Producing segments that can be read and understood clearly.
3. Ensuring every sentence is assigned to a segment, with no omissions and no reordering.

## Primary Objectives
1. Perform a semantic segmentation of the provided text.
2. For each segment, output:
   - A concise subtitle that captures its main idea or subject.
   - The inclusive sentence index range it covers.
3. Output the segmentation in strict JSON format.

## Constraints
1. **Full Coverage and Original Order**: Segment the entire text in its original sentence order. Do not skip any sentence. Every sentence must belong to at least one segment's main index range.
2. **Indexing Rule**: Indices are 1-based and inclusive. The "from_idx" of the first segment must be 1. The "to_idx" of the last segment must equal the last index of sentences in the text. Indices are like [1] in the original text.
3. **Context Indices**: If a segment requires a sentence from outside its main range to be fully comprehensible, you may list that sentence's index in "context_idx". However, a sentence cannot exist *only* in "context_idx"; every sentence must appear as a main member of exactly one segment (i.e., within its "from_idx"-"to_idx" range).
4. **No External Knowledge**: Segment based only on the semantic content of the text itself. Do not use any external knowledge or assumptions about the text.
5. **Semantic Clarity**: Each segment should represent a clear and coherent meaning.
6. **Segment Length**: The recommended maximum length for any single segment is 100 sentences. This is a guidance threshold rather than an absolute hard cap. When necessary, a segment may exceed this limit. Do not artificially merge or split to reach a target length.
7. **Subtitle**: Provide a very short, descriptive subtitle for each segment.

Return STRICT JSON only (no markdown):

{
    "chapter_title": "The chapter title or identifier",
    "segments": [
        {
            "subtitle": "The subtitle of the plot block",
            "from_idx": "Integer of the index where the segment starts, with optional overlap with previous segment if necessary. The first segment must start with index 1.",
            "to_idx": "Integer of the index where the segment ends, with optional overlap with next segment if necessary. The last segment must end with index {last_index}.",
            "context_idx": "List of indices of sentences that provide necessary context for understanding this segment, if necessary."
        }
    ]
}
\end{promptbox}

\begin{promptbox}{System Prompt for Ablation w/o Event Extraction}
# Role: Narratology Segmentation Expert

## Profile:
- Language: English
- Description: You are a professional literary editor and Natural Language Processing expert deeply versed in structuralism and narratology. Your core task is to perform fine-grained plot segmentation on literary texts.

## Core Competencies
1. Fine-grained plot segmentation preserving original literary aesthetics.
2. Untangling complex multi-line, parallel, and nested narrative structures based on "Unity of Action" and internal story logic.
3. Pinpointing logical breakpoints by synthesizing macro-structural shifts (spatial, temporal, tension-based, etc.) with micro-level evaluations of semantic dependency.

## Primary Objectives
1. Perform logical, narratology-driven plot segmentation on literary texts.
2. Enable readers to independently read and understand each segmented snippet out of its full context.
3. Output the exact analytical and segmentation results in strict JSON format.

## Constraints
1. Role Consistency: Don't break character or bypass the narratological framework under any circumstance.
2. Narrative-Driven Granularity: The recommended maximum length for any single segment is 100 sentences. This is a guidance threshold rather than an absolute hard cap. When narrative coherence clearly requires it, a segment may exceed this limit. NEVER unnaturally merge distinct narrative blocks just to hit a sentence count. Abandon any "mathematical division" mindset. Breakpoints must be dictated organically by narratological theories.
3. Conditional Contextual Overlap: A small overlap of key sentences between adjacent plot blocks is allowed ONLY to ensure narrative continuity. This is NOT mandatory. If a natural narrative breakpoint is clear and coherent, make a clean, hard cut. Do not blindly copy previous sentences just for the sake of overlapping.
4. Descriptive Subtitle: Extract a highly concise subtitle that summarizes the core action or reversal event for each plot block.
5. Full Coverage and Order: You must segment the entire chapter in its original order. Do not skip any sentences or paragraphs. Every sentence must belong to at least one segment, i.e., covered in the index range between "from_idx" and "to_idx" of at least one segment.
6. Indexing Rule: The "from_idx" of the first segment MUST be 1. The "to_idx" of the last segment MUST be the last index of the chapter. Indices are 1-based and inclusive like [1] in the original text.
7. Context Indices: For sentences crucial for understanding a segment but outside its main range, include their indices in the "context_idx" field. Sentences cannot ONLY exist in "context_idx"; they must be part of some segment's main range.

## Narratological Workflow
When determining logical breakpoints for plot segmentation, you must execute the following two-step narratological framework sequentially:

### Step 1: Untangle Narrative Threads and Clarify Structures
Identify and extract the narrative threads directly from the provided text using the following criteria:
1. Apply "Unity of Action": Group events by a complete core action, abandoning the "same character" stereotype. Separate unrelated actions performed by the same character; merge joint actions advanced by different characters.
2. Track the "Revelation of Secrets": Use the gradual unfolding of author-planted secrets as a tight internal logic to connect scattered events into a cohesive narrative chain.
3. Clarify Parallel & Minor Threads: Explicitly identify and segment concurrent event chains (e.g., A1 -> B1 -> A2 -> B2). Never merge minor threads into major ones; preserve their structural independence.
4. Identify Nested Hierarchy: Classify the thread's narrative level: [Extradiegetic] (story introduction), [Intradiegetic] (core main plot), or [Metadiegetic] (a story within a story).
5. Establish Basic Data: For each thread, map the timeline, causeline, and critical shift points in time, space, and perspective.

### Step 2: Locate Key Turning Points and Execute Segmentation
Use the data established in Step 1 to locate logical breakpoints. Apply these markers organically, segmenting the text only when shifts clearly and naturally occur:
1. Macro-Structural & Plot Tension: Track tension through the overarching arc (Exposition -> Predicament -> Extrication). Consider segmenting the text at the boundaries or transitions between these major driving events:
   - Opportunity: The introductory event that triggers the story after the background is set.
   - Change of Plans: The event where the main goal is defined and action intensifies.
   - Point of No Return: The event pushing characters to fully commit to their goal.
   - Major Setback & Reversal: The event where the situation completely falls apart, marking a formal reversal of fortune (peripeteia). This generally serves as a strong signal for segmentation.
   - Climax & Recognition: The final resolution and peak tension, often accompanied by a cognitive shift from ignorance to truth (anagnorisis).
   - Aftermath & Buffer: Post-climax events that resolve the predicament and lower tension, smoothly transitioning into the final Extrication phase.
2. Contextual Shifts: Segment at definitive leaps in time, space, or perspective. Cut when the narrative switches between parallel threads or crosses nested narrative levels.
3. Micro-Dynamic Shifts (State & Action): Segment when a situation's fundamental state transitions (equilibrium -> disequilibrium -> new equilibrium) or at the precise moment an action's outcome (success/failure) is revealed.
4. Sentence-Level Boundary Pinpointing: After narrowing down a general turning point, actively evaluate the semantic dependency between adjacent sentences to pinpoint the exact breakpoint:
   - High Semantic Cohesion Zone (Avoid Segmenting): Within the scope of the potential turning point, identify sentences with strong semantic dependency or interlocking logic. Examples include adjacency pairs (e.g., question/answer, attack/defense), continuous pronoun chains, or an immediate physical/emotional reaction to a specific action. Avoid executing a cut within these sequences.
   - Low Semantic Cohesion Zone (Execute the Cut): Within the scope of the potential turning point, locate the exact sentence where the semantic dependency drops to its lowest level. Make the cut right before the sentence that can semantically stand alone--one that initiates a new topic, phase, or action, and requires minimal context from the immediately preceding sentence to be fully understood.

## Output Specifications
1. JSON Format: You must output STRICT JSON format. Escape all internal double quotes within JSON string fields using a backslash.
2. Content Requirement: For each segmented plot block, output a concise subtitle summarizing the core action, and the index ranges based on the constraints.

## Initialization:
As a Narratology Segmentation Expert, you must strictly follow the Constraints and the Narratological Workflow. To initiate the analysis, await the provision of the Chapter Text Content. Upon receipt, execute the two workflow steps internally and output the segmentation results exclusively in the requested STRICT JSON format.

Return STRICT JSON only (no markdown):

{
  "step1": {
    "step_1_threads": "A string briefly describing the narrative threads and their nested hierarchy, based on unity of action, secrets, and key shifts in time, space, or perspective."
  },
  "step2": {
    "step_2_turning_points": [
      {
        "type": "The type of turning point, do not force all predefined types.",
        "description": "A string briefly describing the specific event or shift that constitutes this turning point, explicitly justified by a macro/micro narrative transition or a drop in semantic cohesion."
      }
    ],
    "chapter_title": "The chapter title or identifier",
    "segments": [
      {
        "subtitle": "The subtitle of the plot block",
        "from_idx": "Integer of the index where the segment starts, with optional overlap with previous segment if necessary. The first segment must start with index 1.",
        "to_idx": "Integer of the index where the segment ends, with optional overlap with next segment if necessary. The last segment must end with index {last_index}.",
        "context_idx": "List of indices of sentences that provide necessary context for understanding this segment, if necessary."
      }
    ]
  }
}
\end{promptbox}

\begin{promptbox}{System Prompt for Ablation w/o Thread Untangling}
# Role: Narratology Segmentation Expert

## Profile:
- Language: English
- Description: You are a professional literary editor and Natural Language Processing expert deeply versed in structuralism and narratology. Your core task is to perform fine-grained plot segmentation on literary texts.

## Core Competencies
1. Fine-grained plot segmentation preserving original literary aesthetics.
2. Extracting valid dynamic events and effectively filtering out static or non-narrative noise.
3. Pinpointing logical breakpoints by synthesizing macro-structural shifts (spatial, temporal, tension-based, etc.) with micro-level evaluations of semantic dependency.

## Primary Objectives
1. Perform logical, narratology-driven plot segmentation on literary texts.
2. Enable readers to independently read and understand each segmented snippet out of its full context.
3. Output the exact analytical and segmentation results in strict JSON format.

## Constraints
1. Role Consistency: Don't break character or bypass the narratological framework under any circumstance.
2. Narrative-Driven Granularity: The recommended maximum length for any single segment is 100 sentences. This is a guidance threshold rather than an absolute hard cap. When narrative coherence clearly requires it, a segment may exceed this limit. NEVER unnaturally merge distinct narrative blocks just to hit a sentence count. Abandon any "mathematical division" mindset. Breakpoints must be dictated organically by narratological theories.
3. Conditional Contextual Overlap: A small overlap of key sentences between adjacent plot blocks is allowed ONLY to ensure narrative continuity. This is NOT mandatory. If a natural narrative breakpoint is clear and coherent, make a clean, hard cut. Do not blindly copy previous sentences just for the sake of overlapping.
4. Descriptive Subtitle: Extract a highly concise subtitle that summarizes the core action or reversal event for each plot block.
5. Full Coverage and Order: You must segment the entire chapter in its original order. Do not skip any sentences or paragraphs. Every sentence must belong to at least one segment, i.e., covered in the index range between "from_idx" and "to_idx" of at least one segment.
6. Indexing Rule: The "from_idx" of the first segment MUST be 1. The "to_idx" of the last segment MUST be the last index of the chapter. Indices are 1-based and inclusive like [1] in the original text.
7. Context Indices: For sentences crucial for understanding a segment but outside its main range, include their indices in the "context_idx" field. Sentences cannot ONLY exist in "context_idx"; they must be part of some segment's main range.

## Narratological Workflow
When determining logical breakpoints for plot segmentation, you must execute the following two-step narratological framework sequentially:

### Step 1: Extract Valid Events and Filter Noise
1. Focus on "Dynamic" Changes: Treat the plot as a sequence of dynamic events. Ignore static character portraits or pure scenery descriptions. Extract an event ONLY when characters or objects undergo dynamic changes.
2. Filter by Event Type: Retain "Changes of state" (physical or mental state changes) and "Process events" (actions or happenings without state change, such as talking, thinking, and feeling) as core events. Discard "Stative events" (static physical or mental states) and "Non-events" (generic statements, counterfactuals, questions) to aggressively reduce noise.
3. Handling Ambiguous Paragraphs: For borderline paragraphs, gauge narrativity by checking for: a situation, an agent, one or more sequential actions, a potential object, a spatial location, a temporal specification, and a rationale. A higher density of these features indicates a valid, independent event.

### Step 2: Locate Key Turning Points and Execute Segmentation
Use the valid events extracted in Step 1 to locate logical breakpoints. Apply these markers organically, segmenting the text only when shifts clearly and naturally occur:
1. Macro-Structural & Plot Tension: Track tension through the overarching arc (Exposition -> Predicament -> Extrication). Consider segmenting the text at the boundaries or transitions between these major driving events:
   - Opportunity: The introductory event that triggers the story after the background is set.
   - Change of Plans: The event where the main goal is defined and action intensifies.
   - Point of No Return: The event pushing characters to fully commit to their goal.
   - Major Setback & Reversal: The event where the situation completely falls apart, marking a formal reversal of fortune (peripeteia). This generally serves as a strong signal for segmentation.
   - Climax & Recognition: The final resolution and peak tension, often accompanied by a cognitive shift from ignorance to truth (anagnorisis).
   - Aftermath & Buffer: Post-climax events that resolve the predicament and lower tension, smoothly transitioning into the final Extrication phase.
2. Contextual Shifts: Segment at definitive leaps in time, space, or perspective. Cut when the narrative crosses macro settings.
3. Micro-Dynamic Shifts (State & Action): Segment when a situation's fundamental state transitions (equilibrium -> disequilibrium -> new equilibrium) or at the precise moment an action's outcome (success/failure) is revealed.
4. Sentence-Level Boundary Pinpointing: After narrowing down a general turning point, actively evaluate the semantic dependency between adjacent sentences to pinpoint the exact breakpoint:
   - High Semantic Cohesion Zone (Avoid Segmenting): Within the scope of the potential turning point, identify sentences with strong semantic dependency or interlocking logic. Examples include adjacency pairs (e.g., question/answer, attack/defense), continuous pronoun chains, or an immediate physical/emotional reaction to a specific action. Avoid executing a cut within these sequences.
   - Low Semantic Cohesion Zone (Execute the Cut): Within the scope of the potential turning point, locate the exact sentence where the semantic dependency drops to its lowest level. Make the cut right before the sentence that can semantically stand alone--one that initiates a new topic, phase, or action, and requires minimal context from the immediately preceding sentence to be fully understood.

## Output Specifications
1. JSON Format: You must output STRICT JSON format. Escape all internal double quotes within JSON string fields using a backslash.
2. Content Requirement: For each segmented plot block, output a concise subtitle summarizing the core action, and the index ranges based on the constraints.

## Initialization:
As a Narratology Segmentation Expert, you must strictly follow the Constraints and the Narratological Workflow. To initiate the analysis, await the provision of the Chapter Text Content. Upon receipt, execute the two workflow steps internally and output the segmentation results exclusively in the requested STRICT JSON format.

Return STRICT JSON only (no markdown):

{
  "step1": {
    "step_1_events": "A string briefly listing the core dynamic events extracted (filtering out noise)."
  },
  "step2": {
    "step_2_turning_points": [
      {
        "type": "The type of turning point, do not force all predefined types.",
        "description": "A string briefly describing the specific event or shift that constitutes this turning point, explicitly justified by a macro/micro narrative transition or a drop in semantic cohesion."
      }
    ],
    "chapter_title": "The chapter title or identifier",
    "segments": [
      {
        "subtitle": "The subtitle of the plot block",
        "from_idx": "Integer of the index where the segment starts, with optional overlap with previous segment if necessary. The first segment must start with index 1.",
        "to_idx": "Integer of the index where the segment ends, with optional overlap with next segment if necessary. The last segment must end with index {last_index}.",
        "context_idx": "List of indices of sentences that provide necessary context for understanding this segment, if necessary."
      }
    ]
  }
}
\end{promptbox}

\begin{promptbox}{System Prompt for Ablation w/o Turning Points Pinpointing}
# Role: Narratology Segmentation Expert

## Profile:
- Language: English
- Description: You are a professional literary editor and Natural Language Processing expert deeply versed in structuralism and narratology. Your core task is to perform fine-grained plot segmentation on literary texts.

## Core Competencies
1. Fine-grained plot segmentation preserving original literary aesthetics.
2. Extracting valid dynamic events and effectively filtering out static or non-narrative noise.
3. Untangling complex multi-line, parallel, and nested narrative structures based on "Unity of Action" and internal story logic.
4. Executing text segmentation to establish clear start and end sentence boundaries for each narrative block.


## Primary Objectives
1. Perform logical, narratology-driven plot segmentation on literary texts.
2. Enable readers to independently read and understand each segmented snippet out of its full context.
3. Output the exact analytical and segmentation results in strict JSON format.

## Constraints
1. Role Consistency: Don't break character or bypass the narratological framework under any circumstance.
2. Narrative-Driven Granularity: The recommended maximum length for any single segment is 100 sentences. This is a guidance threshold rather than an absolute hard cap. When narrative coherence clearly requires it, a segment may exceed this limit. NEVER unnaturally merge distinct narrative blocks just to hit a sentence count. Abandon any "mathematical division" mindset. Breakpoints must be dictated organically by narratological theories.
3. Conditional Contextual Overlap: A small overlap of key sentences between adjacent plot blocks is allowed ONLY to ensure narrative continuity. This is NOT mandatory. If a natural narrative breakpoint is clear and coherent, make a clean, hard cut. Do not blindly copy previous sentences just for the sake of overlapping.
4. Descriptive Subtitle: Extract a highly concise subtitle that summarizes the core action or reversal event for each plot block.
5. Full Coverage and Order: You must segment the entire chapter in its original order. Do not skip any sentences or paragraphs. Every sentence must belong to at least one segment, i.e., covered in the index range between "from_idx" and "to_idx" of at least one segment.
6. Indexing Rule: The "from_idx" of the first segment MUST be 1. The "to_idx" of the last segment MUST be the last index of the chapter. Indices are 1-based and inclusive like [1] in the original text.
7. Context Indices: For sentences crucial for understanding a segment but outside its main range, include their indices in the "context_idx" field. Sentences cannot ONLY exist in "context_idx"; they must be part of some segment's main range.

## Narratological Workflow
When determining logical breakpoints for plot segmentation, you must execute the following three-step narratological framework sequentially:

### Step 1: Extract Valid Events and Filter Noise
1. Focus on "Dynamic" Changes: Treat the plot as a sequence of dynamic events. Ignore static character portraits or pure scenery descriptions. Extract an event ONLY when characters or objects undergo dynamic changes.
2. Filter by Event Type: Retain "Changes of state" (physical or mental state changes) and "Process events" (actions or happenings without state change, such as talking, thinking, and feeling) as core events. Discard "Stative events" (static physical or mental states) and "Non-events" (generic statements, counterfactuals, questions) to aggressively reduce noise.
3. Handling Ambiguous Paragraphs: For borderline paragraphs, gauge narrativity by checking for: a situation, an agent, one or more sequential actions, a potential object, a spatial location, a temporal specification, and a rationale. A higher density of these features indicates a valid, independent event.

### Step 2: Untangle Narrative Threads and Clarify Structures
Group the events extracted in Step 1 into clear narrative threads using the following criteria:
1. Apply "Unity of Action": Group events by a complete core action, abandoning the "same character" stereotype. Separate unrelated actions performed by the same character; merge joint actions advanced by different characters.
2. Track the "Revelation of Secrets": Use the gradual unfolding of author-planted secrets as a tight internal logic to connect scattered events into a cohesive narrative chain.
3. Clarify Parallel & Minor Threads: Explicitly identify and segment concurrent event chains (e.g., A1 -> B1 -> A2 -> B2). Never merge minor threads into major ones; preserve their structural independence.
4. Identify Nested Hierarchy: Classify the thread's narrative level: [Extradiegetic] (story introduction), [Intradiegetic] (core main plot), or [Metadiegetic] (a story within a story).
5. Establish Basic Data: For each thread, map the timeline, causeline, and critical shift points in time, space, and perspective.

### Step 3: Execute Segmentation
Use the data established in Step1 and Step 2 to directly execute the text segmentation.

## Output Specifications
1. JSON Format: You must output STRICT JSON format. Escape all internal double quotes within JSON string fields using a backslash.
2. Content Requirement: For each segmented plot block, output a concise subtitle summarizing the core action, and the index ranges based on the constraints.

## Initialization:
As a Narratology Segmentation Expert, you must strictly follow the Constraints and the Narratological Workflow. To initiate the analysis, await the provision of the Chapter Text Content. Upon receipt, execute all three workflow steps internally and output the segmentation results exclusively in the requested STRICT JSON format.

Return STRICT JSON only (no markdown):

{
  "step1": {
    "step_1_events": "A string briefly listing the core dynamic events extracted (filtering out noise)."
  },
  "step2": {
    "step_2_threads": "A string briefly describing the narrative threads and their nested hierarchy, based on unity of action, secrets, and key shifts in time, space, or perspective."
  },
  "step3": {
    "chapter_title": "The chapter title or identifier",
    "segments": [
      {
        "subtitle": "The subtitle of the plot block",
        "from_idx": "Integer of the index where the segment starts, with optional overlap with previous segment if necessary. The first segment must start with index 1.",
        "to_idx": "Integer of the index where the segment ends, with optional overlap with next segment if necessary. The last segment must end with index {last_index}.",
        "context_idx": "List of indices of sentences that provide necessary context for understanding this segment, if necessary."
      }
    ]
  }
}
\end{promptbox}

\section{Evaluation Metric Overview}
\label{app:metric-details}

\subsection{Exact Match}
\label{app:metric-em}

Exact Match (EM) measures whether a prediction exactly matches a reference answer after normalization. Let $\mathrm{norm}(\cdot)$ denote lowercasing and punctuation-stripping normalization. For a prediction $\hat{a}$ and a set of references $\mathcal{A}$, EM is defined as
\begin{equation}
\mathrm{EM}(\hat{a}, \mathcal{A})
=
\max_{a \in \mathcal{A}}
\mathbb{I}\!\left[
\mathrm{norm}(\hat{a}) =
\mathrm{norm}(a)
\right].
\end{equation}
The final EM score is the average over all examples.

\subsection{F1}
\label{app:metric-f1}

F1 computes token-level overlap between the prediction and the reference answer. For a prediction $\hat{a}$ and a reference $a$, let $T(\hat{a})$ and $T(a)$ denote their normalized token multisets, and let $m$ be the number of overlapping tokens. Precision and recall are
\begin{equation}
P = \frac{m}{|T(\hat{a})|},
\qquad
R = \frac{m}{|T(a)|}.
\end{equation}
The token-level F1 score is
\begin{equation}
\mathrm{F1} =
\frac{2PR}{P+R}.
\end{equation}
When multiple references are available, we use the maximum F1 over references.

\subsection{ROUGE-L}
\label{app:metric-rouge-l}

ROUGE-L~\cite{lin-2004-rouge} measures longest-common-subsequence overlap. Let $\mathrm{LCS}(\hat{a},a)$ be the length of the longest common subsequence between the prediction $\hat{a}$ and a reference answer $a$. ROUGE-L precision and recall are
\begin{equation}
P_{\mathrm{LCS}}
=
\frac{\mathrm{LCS}(\hat{a},a)}{|\hat{a}|},
\qquad
R_{\mathrm{LCS}}
=
\frac{\mathrm{LCS}(\hat{a},a)}{|a|}.
\end{equation}
The ROUGE-L F-score is
\begin{equation}
\mathrm{ROUGE\mbox{-}L}
=
\frac{(1+\beta^2)P_{\mathrm{LCS}}R_{\mathrm{LCS}}}
{R_{\mathrm{LCS}}+\beta^2P_{\mathrm{LCS}}},
\end{equation}
where $\beta$ controls the relative weight of recall.

\subsection{METEOR}
\label{app:metric-meteor}

METEOR~\cite{banerjee-lavie-2005-meteor} aligns unigrams between the prediction and the reference using exact, stem, and synonym matches. Let $P_m$ and $R_m$ denote unigram precision and recall after alignment. METEOR first computes a weighted harmonic mean,
\begin{equation}
F_{\mathrm{mean}}
=
\frac{10P_mR_m}{R_m + 9P_m},
\end{equation}
and then applies a fragmentation penalty,
\begin{equation}
\mathrm{Penalty}
=
\gamma
\left(
\frac{c}{m}
\right)^{\theta},
\end{equation}
where $m$ is the number of matched unigrams and $c$ is the number of contiguous matched chunks. The final score is
\begin{equation}
\mathrm{METEOR}
=
(1-\mathrm{Penalty})F_{\mathrm{mean}}.
\end{equation}

\subsection{Context Relevance}
\label{app:metric-context-relevance}

Context Relevance\footnote{\label{fn:ragas}Available at: \url{https://docs.ragas.io/en/stable/concepts/metrics/available_metrics/nvidia_metrics/}} evaluates whether the retrieved contexts are pertinent to the user input. This is done via two independent LLM-as-a-Judge prompts that each rate the relevance on a scale of 0, 1, or 2. The ratings are then converted to a $[0,1]$ scale and averaged to produce the final score. Given a question $q$ and retrieved contexts $c$, each judge assigns a score $r_j \in \{0,1,2\}$, where 0 means the retrieved contexts are not relevant to the query at all, 1 means the contexts are partially relevant, and 2 means the contexts are completely relevant. The normalized context relevance score for one example is
\begin{equation}
\mathrm{CtxRel}(q,c)
=
\frac{1}{2}
\sum_{j=1}^{2}
\frac{r_j}{2}.
\end{equation}
Higher scores indicate that the contexts are more closely aligned with the user's query. The final score is averaged over examples.

\subsection{Answer Accuracy}
\label{app:metric-answer-accuracy}

Answer Accuracy\footref{fn:ragas} measures the agreement between a model's response and a reference ground truth for a given question. This is done via two distinct LLM-as-a-Judge prompts that each return a rating (0, 2, or 4). The metric converts these ratings into a $[0,1]$ scale and then takes the average of the two scores from the judges. Given a question $q$, prediction $\hat{a}$, and reference answer $a$, each judge assigns a score $s_j \in \{0,2,4\}$, where 0 means the response is inaccurate or does not address the same question as the reference, 2 means the response partially aligns with the reference, and 4 means the response exactly aligns with the reference. The normalized answer accuracy score for one example is
\begin{equation}
\mathrm{AnsAcc}(q,\hat{a},a)
=
\frac{1}{2}
\sum_{j=1}^{2}
\frac{s_j}{4}.
\end{equation}
The two prompts use distinct templates to ensure robustness: one compares the response with the reference directly, while the other swaps the roles of the response and reference. If both ratings are valid, the final score is the average; otherwise, it takes the valid one. Higher scores indicate that the model's answer closely matches the reference. The final score is averaged over examples.

\subsection{Mean Reciprocal Rank}
\label{app:metric-mrr}

For GutenQA, we additionally report Mean Reciprocal Rank (MRR) as a retrieval metric. MRR evaluates the ranking quality based on the gold text span. Let $g_i$ be the gold span for example $i$, and let $d_{i,j}$ be the $j$-th chunk in the reranked list $\mathcal{D}_i$ sorted by reranker score. We define the rank of the first relevant chunk as:
\begin{equation}
\mathrm{rank}_i = \min \{ j \mid g_i \subseteq d_{i,j} \},
\end{equation}
where $\mathrm{rank}_i = \infty$ (and thus $\frac{1}{\mathrm{rank}_i} = 0$) if no chunk in the list contains the gold span. MRR is then calculated as the average of the reciprocal ranks across the $N$ examples:
\begin{equation}
\mathrm{MRR} = \frac{1}{N} \sum_{i=1}^{N} \frac{1}{\mathrm{rank}_i}.
\end{equation}
Before matching, both the gold span and retrieved chunks are lowercased and stripped of whitespace and newline characters. MRR measures how effectively the system places relevant evidence at the top of the ranked list, heavily penalizing systems that require scanning many chunks.

\subsection{Hit@$k$}
\label{app:metric-hitk}

For GutenQA, we additionally report Hit@$k$ as a retrieval metric. Each question is associated with a gold text span that must be contained in the retrieved evidence. We sort reranked chunks by reranker score and check whether the gold span appears in the top-$k$ chunks. Let $g_i$ be the gold span for example $i$, and let $\mathcal{D}_{i,k}$ be the top-$k$ reranked chunks. Hit@$k$ is
\begin{equation}
\mathrm{Hit@}k
=
\frac{1}{N}
\sum_{i=1}^{N}
\mathbb{I}
\left[
\exists d \in \mathcal{D}_{i,k}
\ \mathrm{s.t.}\ 
g_i \subseteq d
\right].
\end{equation}
Before matching, both the gold span and retrieved chunks are lowercased and stripped of whitespace and newline characters. We report Hit@1, Hit@2, Hit@3, Hit@5, and Hit@20. Hit@$k$ measures whether required evidence is present.

\section{Additional Quantitative Results}
\label{app:gutenqa_automated}

See Tables \ref{tab:gutenqa_automated_baselines} and \ref{tab:gutenqa_automated_ablation}.

\begin{table}[htbp]
\centering
\scriptsize
\resizebox{\columnwidth}{!}{
\begin{tabular}{lccccc}
\toprule
\textbf{Method} & \multicolumn{5}{c}{\textbf{GutenQA}} \\
\cmidrule(lr){2-6}
 & \textbf{EM} & \textbf{F1} & \textbf{R-L} & \textbf{Mtr} & \textbf{AA} \\
\midrule
\textit{RAG Baselines} & & & & & \\
Token & 0.0427 & 0.5389 & 0.5018 & 0.5592 & 0.6396 \\
Recursive Char & 0.0420 & 0.5371 & 0.4999 & 0.5584 & 0.6246 \\
Perplexity & 0.0390 & 0.5327 & 0.4962 & 0.5535 & 0.6129 \\
\quad + Merge & 0.0383 & 0.5340 & 0.4963 & 0.5570 & 0.6268 \\
Margin-Sampling & 0.0257 & 0.4959 & 0.4640 & 0.5103 & 0.5154 \\
\quad + Merge & 0.0327 & 0.5255 & 0.4896 & 0.5492 & 0.6128 \\
LumberChunker & 0.0343 & 0.5276 & 0.4887 & 0.5538 & 0.6229 \\
\quad + Merge & 0.0380 & 0.5256 & 0.4889 & 0.5527 & 0.6262 \\
\midrule
\textit{Ours} & & & & & \\
LitSeg & 0.0320 & 0.5174 & 0.4791 & 0.5525 & 0.6252 \\
\bottomrule
\end{tabular}
}
\caption{Automated evaluation of generation results on GutenQA.}
\label{tab:gutenqa_automated_baselines}
\end{table}

\begin{table}[htbp]
\centering
\scriptsize
\setlength{\tabcolsep}{4pt}
\begin{tabular}{lccccc}
\toprule
\textbf{Method} & \multicolumn{5}{c}{\textbf{GutenQA}} \\
\cmidrule(lr){2-6}
& \textbf{EM} & \textbf{F1} & \textbf{R-L} & \textbf{Mtr} & \textbf{AA} \\
\midrule
LitSeg               & 0.0320          & 0.5174          & 0.4791          & 0.5525 & 0.6252          \\
\hspace{0.5em}w/o GRPO       & 0.0320          & 0.5159          & 0.4780          & 0.5521 & 0.6252          \\
\hspace{0.5em}w/o FT        & 0.0323 & 0.5191          & 0.4801          & 0.5519          & 0.6259          \\
\hspace{0.5em}+w/o Theo.S.1 & 0.0273          & 0.5107          & 0.4716          & 0.5447          & 0.6200          \\
\hspace{0.5em}+w/o Theo.S.2 & 0.0317          & 0.5154          & 0.4752          & 0.5479          & 0.6286 \\
\hspace{0.5em}+w/o Theo.S.3 & 0.0310          & 0.5154          & 0.4775          & 0.5458          & 0.6195          \\
\hspace{0.5em}+w/o Theory    & 0.0307          & 0.5202 & 0.4830 & 0.5511          & 0.6262          \\
\bottomrule
\end{tabular}
\caption{Automated evaluation of ablation study on GutenQA.}
\label{tab:gutenqa_automated_ablation}
\end{table}

\section{Baseline Overview}
\label{app:baseline-overview}

\paragraph{Token splitter.}
The token splitter divides text into fixed-length token windows with overlap. It is a length-based baseline and does not model semantics.

\paragraph{Recursive-character splitter.}
The recursive-character splitter creates chunks according to a target character length. It recursively chunks and falls back to smaller separators, including paragraph breaks, newlines, punctuation, and whitespace.

\paragraph{Perplexity chunker.}
Perplexity Chunker~\cite{zhao2025metachunkinglearningtextsegmentation} computes language-model loss for each sentence and identifies local minima in the sentence-level loss curve as potential boundaries. Let $s_i$ denote the $i$-th sentence. We compute the average token-level cross-entropy loss:
\begin{equation}
\mathcal{L}(s_i)
=
\frac{1}{|s_i|}
\sum_{t \in s_i}
\mathrm{CE}\left(p_{\theta}(t \mid t_{<i}), t\right),
\end{equation}
where $p_{\theta}$ is the language-model distribution. A sentence is selected as a boundary if it is a local minimum and the loss drop is sufficiently large:
\begin{equation}
\begin{aligned}
&\mathcal{L}(s_i) < \mathcal{L}(s_{i-1})
\ \wedge\
\mathcal{L}(s_i) < \mathcal{L}(s_{i+1}) \\
&\quad \wedge\
\left(
\Delta_{i-1} \ge \tau
\ \vee\
\Delta_{i+1} \ge \tau
\right),
\end{aligned}
\end{equation}
where $\Delta_j = \mathcal{L}(s_j) - \mathcal{L}(s_i)$ and $\tau=0.5$ by default. In implementation, the selected minimum sentence is included in both neighboring chunks, which creates a one-sentence overlap at the detected boundary. This method is sensitive to distributional or stylistic changes but does not explicitly encode narratological criteria.

\paragraph{Margin-sampling chunker.}
Margin-Sampling Chunker~\cite{zhao2025metachunkinglearningtextsegmentation} treats each candidate boundary as a binary decision between splitting and keeping adjacent text together. Given the current chunk $L$ and the next sentence $R$, the model is prompted to choose between option ``1'' for split and option ``2'' for keep.

\begin{promptbox}{Margin-sampling decision prompt}
This is a text chunking task. You are a text analysis expert. Please group two related paragraphs together and separate unrelated paragraphs based on the logical structure and semantic content of the provided sentences. Choose one chunking method from the following two options according to the above requirements:
1. Split "{left} {right}" into "{left}" and "{right}";
2. Keep "{left} {right}" unsplit in its original form;
Please answer 1 or 2.
\end{promptbox}

Let $p_1$ and $p_2$ denote the model probabilities assigned to option ``1'' and option ``2'', respectively. The boundary score is
\begin{equation}
\mathrm{score}(L,R) = p_2 - p_1.
\end{equation}
The chunker keeps $L$ and $R$ together when $\mathrm{score}(L,R) > \theta$ and splits otherwise. The threshold $\theta$ is initialized to 0 and updated as the moving average of the most recent five scores:
\begin{equation}
\theta_t
=
\frac{1}{|\mathcal{H}_t|}
\sum_{u \in \mathcal{H}_t} u,
\qquad
|\mathcal{H}_t| \le 5,
\end{equation}
where $\mathcal{H}_t$ contains the most recent margin scores.

\paragraph{LumberChunker.}
LumberChunker~\cite{duarte-etal-2024-lumberchunker} follows the idea of prompting an LLM to identify content shifts. A sliding window of sentence-level units is formatted as a numbered list and presented to the model as paragraphs. The model is asked to return the first ID whose content clearly changes compared with previous units.

\begin{promptbox}{LumberChunker prompt}
You will receive as input an English document with paragraphs identified by 'ID XXXX: <text>'.
Task: Find the first paragraph (not the first one) where the content clearly changes compared to the previous paragraphs.
Output: Return the ID of the paragraph with the content shift as in the exemplified format: 'Answer: ID XXXX'.
Additional Considerations: Avoid very long groups of paragraphs. Aim for a good balance between identifying content shifts and keeping groups manageable.
Document:
{numbered_units}
\end{promptbox}

Let $i_t$ be the current start position and let $W_t=\{u_{i_t},\ldots,u_{j_t}\}$ be the current window, whose total length is bounded by 550 words. The model returns an index $\hat{b}_t$ inside the window, and the next chunk is
\begin{equation}
C_t = [u_{i_t}, \ldots, u_{i_t+\hat{b}_t-1}].
\end{equation}
The start position is then updated to $i_{t+1}=i_t+\hat{b}_t$. Invalid responses are retried up to a predefined number of times. If all retries fail, the chunker falls back to splitting after the first unit in the current window.

\paragraph{Dynamic merge in baselines.}
For perplexity, margin-sampling, and LumberChunker, we also evaluate a dynamic-merge variant. Dynamic merge greedily merges adjacent chunks when their combined length is no more than 200 words. Given chunks $C_i$ and $C_{i+1}$, the merge condition is
\begin{equation}
|C_i|_{\mathrm{word}} + |C_{i+1}|_{\mathrm{word}}
\le 200.
\end{equation}
This post-processing step reduces extremely short chunks that may be harmful for retrieval.

\section{Implementation Details}
\label{app:impl}

This appendix describes our implementation details.

\subsection{RAG Pipeline}
\label{app:rag-pipeline}

\subsubsection{Pipeline Overview}
\label{app:rag-overview}

Our RAG system follows a classic yet robust retrieve--rerank--generate architecture. Specifically, the pipeline consists of the following stages:
\begin{enumerate}
    \item \textbf{Text loading.} 
    \item \textbf{Text splitting.} The target chunker segments each chapter into chunks.
    \item \textbf{Dense indexing.} Chunks are encoded by a fixed embedding model and stored in vector databases.
    \item \textbf{Hybrid retrieval.} Dense similarity search and sparse search retrieve candidate chunks under book-level metadata filters.
    \item \textbf{Reranking.} A fixed reranker scores the retrieved candidates and selects the top chunks.
    \item \textbf{Answer generation.} A fixed generator produces a concise answer based on the reranked evidence.
\end{enumerate}

\subsubsection{RAG Implementation Details}
\label{app:rag-details}

Table~\ref{tab:rag-config} summarizes the RAG configuration. All components are shared and kept unchanged across chunkers. See Table \ref{tab:generator-hyperparams} for generator configuration.

\begin{table}[htbp]
\centering
\scriptsize
\setlength{\tabcolsep}{3pt}
\renewcommand{\arraystretch}{1.06}
\begin{tabular}{@{}p{0.37\columnwidth}p{0.55\columnwidth}@{}}
\toprule
\textbf{Component} & \textbf{Configuration} \\
\midrule
Variable component & Text splitter (chunker) \\
Vector store & Chroma \\
Embedding model & Qwen/Qwen3-Embedding-8B \\
Dense retrieval & Similarity search \\
Dense top-$k$ & 18 \\
Sparse retrieval & BM25 \\
Sparse top-$k$ & 2 \\
Candidate fusion & Union of dense and sparse candidates \\
Reranker & Qwen/Qwen3-Reranker-8B \\
Reranker top-$k$ & 5 \\
Generator & Qwen/Qwen3.5-9B \\
Generator mode & Non-thinking mode \\
\bottomrule
\end{tabular}
\caption{RAG pipeline configuration.}
\label{tab:rag-config}
\end{table}

\begin{table}[htbp]
\centering
\scriptsize
\setlength{\tabcolsep}{3pt}
\renewcommand{\arraystretch}{1.06}
\begin{tabular}{@{}p{0.43\columnwidth}p{0.49\columnwidth}@{}}
\toprule
\textbf{Parameter} & \textbf{Value} \\
\midrule
Model & Qwen/Qwen3.5-9B \\
Thinking mode & Disabled \\
Temperature & 1.0 \\
TopP & 0.95 \\
TopK & 20 \\ 
MinP & 0.0 \\
Presence Penalty & 1.5 \\
Repetition Penalty & 1.0 \\
\bottomrule
\end{tabular}
\caption{RAG pipeline generator configuration.}
\label{tab:generator-hyperparams}
\end{table}

\begin{promptbox}{Embedding and query instruction}
Represent the literature text chunk for literary question answering:
\end{promptbox}

\begin{promptbox}{Reranker instruction}
Given a literary question, retrieve relevant literature text chunks that answer the question.
\end{promptbox}

\begin{promptbox}{Generation system prompt}
You are a helpful and precise assistant for answering questions about literature works. Your task is to answer the question based on information provided. If you cannot find the answer in the provided information, say you don't know. Answer directly to the question without explanation or additional information.
\end{promptbox}

\begin{promptbox}{Generation system prompt benchmark-specific suffixes}
LiteraryQA:
Answer in one phrase or one sentence, as concise as possible.

GutenQA:
Answer in one sentence, and be concise.
\end{promptbox}

\begin{promptbox}{Non-RAG generation user prompt template}
Book Title: {title}

Question: {query}
\end{promptbox}

\begin{promptbox}{RAG generation user prompt template}
Book Title: {title}

Question: {query}

References:
Chunk:
{doc}
\end{promptbox}

\subsection{LitSeg Chunker}
\label{app:teacher-data-generation}

See Table \ref{tab:teacher-hyperparams}.

\begin{table}[htbp]
\centering
\scriptsize
\setlength{\tabcolsep}{3pt}
\renewcommand{\arraystretch}{1.06}
\begin{tabular}{@{}p{0.43\columnwidth}p{0.49\columnwidth}@{}}
\toprule
\textbf{Parameter} & \textbf{Value} \\
\midrule
Model & Qwen/Qwen3.6-27B-FP8 \\
Sentence tokenizer & NLTK~\cite{bird-loper-2004-nltk} \\
Input chunk size & 25,000 words \\
Maximum retries & 20 \\
Thinking mode & Disabled \\
Temperature & 1.0 \\
TopP & 0.95 \\
TopK & 20 \\ 
MinP & 0.0 \\
Presence Penalty & 1.5 \\
Repetition Penalty & 1.0 \\
\bottomrule
\end{tabular}
\caption{LitSeg chunker configuration.}
\label{tab:teacher-hyperparams}
\end{table}

\subsection{LitSeg-Lite Chunker}
\subsubsection{Inference Details}
\label{app:litseg-lite-inf-impl}

See Table \ref{tab:litseg-infer-config}.

\begin{table}[htbp]
\centering
\scriptsize
\setlength{\tabcolsep}{3pt}
\renewcommand{\arraystretch}{1.06}
\begin{tabular}{@{}p{0.40\columnwidth}p{0.52\columnwidth}@{}}
\toprule
\textbf{Parameter} & \textbf{Value} \\
\midrule
Base model & Qwen/Qwen3-4B-Instruct-2507 \\
Sentence tokenizer & NLTK \\
Input chunk size & 25,000 words \\
Maximum retries & 20 \\
Temperature & 0.7 \\
TopP & 0.8 \\
TopK & 20 \\ 
MinP & 0 \\
\bottomrule
\end{tabular}
\caption{LitSeg-Lite chunker configuration.}
\label{tab:litseg-infer-config}
\end{table}

\subsubsection{Training Details}
\label{app:litseg-lite-training-details}

See Table \ref{tab:sft-hyperparams}, \ref{tab:grpo-hyperparams}, \ref{tab:grpo-rm-config} and \ref{tab:litseg-lite-format-reward}.

\begin{table}[htbp]
\centering
\scriptsize
\setlength{\tabcolsep}{3pt}
\renewcommand{\arraystretch}{1.06}
\begin{tabular}{@{}p{0.43\columnwidth}p{0.49\columnwidth}@{}}
\toprule
\textbf{Parameter} & \textbf{Value} \\
\midrule
Base model & Qwen/Qwen3-4B-Instruct-2507 \\
Precision & fp16 \\
PEFT method & LoRA \\
LoRA rank $r$ & 16 \\
LoRA alpha $\alpha$ & 32 \\
LoRA dropout & 0 \\
LoRA bias & none \\
Target modules & q, k, v, o, gate, up, down projections \\
Epochs & 3 \\
Learning rate & $2\times10^{-4}$ \\
Per-device batch size & 1 \\
Gradient accumulation & 8 \\
Optimizer & adamw\_torch\_fused \\
Random seed & 42 \\
\bottomrule
\end{tabular}
\caption{LitSeg-Lite SFT configuration.}
\label{tab:sft-hyperparams}
\end{table}

\begin{table}[htbp]
\centering
\scriptsize
\setlength{\tabcolsep}{3pt}
\renewcommand{\arraystretch}{1.06}
\begin{tabular}{@{}p{0.43\columnwidth}p{0.49\columnwidth}@{}}
\toprule
\textbf{Parameter} & \textbf{Value} \\
\midrule
Base model & Qwen/Qwen3-4B-Instruct-2507 \\
Loss type & DAPO \\
Reward weights & Format: 0.1, Model-based: 0.9 \\
Generations per prompt ($G$) & 8 \\
Iterations per batch & 4 \\
Epochs & 2 \\
Learning rate & $5\times10^{-6}$ \\
Per-device batch size & 1 \\
Gradient accumulation & 8 \\
Sampling temperature & 0.7 \\
Optimizer & adamw\_torch\_fused \\
Random seed & 42 \\
\bottomrule
\end{tabular}
\caption{LitSeg-Lite GRPO configuration.}
\label{tab:grpo-hyperparams}
\end{table}

\begin{table}[htbp]
\centering
\scriptsize
\setlength{\tabcolsep}{3pt}
\renewcommand{\arraystretch}{1.06}
\begin{tabular}{@{}p{0.43\columnwidth}p{0.49\columnwidth}@{}}
\toprule
\textbf{Parameter} & \textbf{Value} \\
\midrule
Model & Qwen/Qwen3.6-27B-FP8 \\
Thinking mode & Enabled \\
Temperature & 1.0 \\
TopP & 0.95 \\
TopK & 20 \\ 
MinP & 0.0 \\
Presence Penalty & 1.5 \\
Repetition Penalty & 1.0 \\
\bottomrule
\end{tabular}
\caption{LitSeg-Lite GRPO reward model configuration.}
\label{tab:grpo-rm-config}
\end{table}

\begin{table}[htbp]
\centering
\scriptsize
\setlength{\tabcolsep}{3pt}
\renewcommand{\arraystretch}{1.06}
\begin{tabular}{@{}p{0.32\columnwidth}p{0.10\columnwidth}p{0.50\columnwidth}@{}}
\toprule
\textbf{Condition} & \textbf{Reward} & \textbf{Description} \\
\midrule
Valid output & $0.0$ & The output is parsable JSON, contains all required fields, and passes segment validation. \\
Invalid JSON & $-1.0$ & The output cannot be parsed as JSON after basic cleanup. \\
Missing required fields & $-0.8$ & The output JSON misses any of \texttt{step1}, \texttt{step2}, or \texttt{step3}. \\
Invalid \texttt{step3} segments & $-0.6$ & The \texttt{step3} segmentation fails validation, e.g., invalid index ranges, invalid context indices, or sentence coverage violations. \\
\bottomrule
\end{tabular}
\caption{Rule-based format reward for LitSeg-Lite GRPO training.}
\label{tab:litseg-lite-format-reward}
\end{table}

\subsection{Evaluation Metrics}
\label{app:metrics-impl}

See Table \ref{tab:ragas-config}. %

\begin{table}[htbp]
\centering
\scriptsize
\setlength{\tabcolsep}{3pt}
\renewcommand{\arraystretch}{1.06}
\begin{tabular}{@{}p{0.43\columnwidth}p{0.49\columnwidth}@{}}
\toprule
\textbf{Parameter} & \textbf{Value} \\
\midrule
Judge model & Qwen/Qwen3.5-9B \\
Thinking mode & Disabled \\
Temperature & 0.0 \\
Seed & 42 \\
\bottomrule
\end{tabular}
\caption{RAGAS metrics configuration.}
\label{tab:ragas-config}
\end{table}

\subsection{Baselines}
\label{app:baseline-details}

See Table~\ref{tab:baseline-config}.

\begin{table}[htbp]
\centering
\scriptsize
\setlength{\tabcolsep}{3pt}
\renewcommand{\arraystretch}{1.06}
\begin{tabular}{@{}
>{\raggedright\arraybackslash}p{0.26\columnwidth}
>{\raggedright\arraybackslash}p{0.70\columnwidth}
@{}}
\toprule
\textbf{Baseline} & \textbf{Configuration} \\
\midrule
Token splitter
& 200 tokens; 40-token overlap. \\
Recursive-character splitter
& 1,000 characters; 200-character overlap. Separators: \texttt{['\textbackslash n', '.', ',', ';', ':', '~']}. \\
Perplexity chunker\tablefootnote{For this baseline, a smaller model is used instead of \texttt{Qwen/Qwen3-4B-Instruct-2507} to avoid resource issues. The perplexity calculation requires caching all intermediate hidden states, a process that is both memory-intensive ($\sim$150\,GB VRAM with the larger model) and incompatible with standard high-throughput inference frameworks.}
& Threshold $\tau=0.5$; selected minimum sentence is shared by adjacent chunks. Model: \texttt{Qwen/Qwen3.5-0.8B}; maximum token window 9000. \\
Perplexity + Merge
& Perplexity chunker followed by 200-word dynamic merge. Model: same as the perplexity chunker. \\
Margin-sampling chunker
& Model: \texttt{Qwen/Qwen3-4B-Instruct-2507}; threshold: initialized to 0 and updated by the latest five scores.  \\
Margin sampling + Merge
& Margin-sampling chunker followed by 200-word dynamic merge. Model: same as the margin-sampling chunker. \\
LumberChunker
& Model: \texttt{Qwen/Qwen3-4B-Instruct-2507}; sliding window size: 550 words; maximum retries 20; invalid outputs fall back to splitting after the first unit. \\
LumberChunker + Merge
& LumberChunker followed by 200-word dynamic merge. Model: same as LumberChunker. \\
\bottomrule
\end{tabular}
\caption{Baseline chunker configuration.}
\label{tab:baseline-config}
\end{table}

\end{document}